\documentclass[nohyperref]{article}

\usepackage{natbib}
\usepackage[top=1in, bottom=1in, left=1in, right=1in]{geometry}
\usepackage[english]{babel}
\usepackage{etoc}
\usepackage{microtype}
\usepackage{graphicx}
\usepackage{xfrac}
\usepackage{subcaption}
\usepackage{booktabs}
\usepackage{multirow}
\usepackage{pifont}
\usepackage{wrapfig}
\usepackage{hyperref}
\usepackage{algorithm}
\usepackage{forloop}
\usepackage{tablefootnote}
\usepackage{algpseudocode}
\usepackage{enumitem}
\usepackage{amsmath}
\usepackage{amssymb}
\usepackage{mathtools}
\usepackage{amssymb}
\usepackage{pifont}
\usepackage{amsthm}
\usepackage[capitalize,noabbrev]{cleveref}
\usepackage{listings}
\usepackage{xcolor}
\definecolor{dkgreen}{rgb}{0,0.6,0}
\definecolor{dkred}{rgb}{0.8,0.0,0}
\definecolor{dkblue}{rgb}{0.0,0.0,0.9}
\definecolor{gray}{rgb}{0.5,0.5,0.5}
\definecolor{mauve}{rgb}{0.58,0,0.82}
\definecolor{lightgray}{HTML}{EEEEEE}
\definecolor{dkcyan}{HTML}{008b8b}

\usepackage[font=small,labelfont=bf]{caption}
\usepackage{tikz}
\usepackage{wrapfig}
\usepackage{thm-restate}

\hypersetup{
    colorlinks=true,
    linkcolor=dkcyan,
    citecolor=dkcyan,
    filecolor=dkcyan,
    urlcolor=dkcyan}
    
\everypar{\looseness=-1}
\setlist[itemize]{leftmargin=*}
\newif\ifcomments

\commentstrue

\ifcomments
  \newcommand{\colornote}[3]{{\color{#1}\bf{(#2) #3}\normalfont}}
\else
  \newcommand{\colornote}[3]{}
\fi

\DeclareMathOperator*{\argmin}{arg\,min}

\newcommand{\ts}{\mathcal{T}}

\newcommand{\is}{\mathcal{X}}

\newcommand{\trainset}{\mathcal{D}_{\text{train}}}
\newcommand{\inp}{\mathbf{x}}
\newcommand{\target}{\mathbf{t}}
\newcommand{\point}{\mathbf{z}}
\newcommand{\out}{\mathbf{y}}

\newcommand{\ucost}{\mathcal{Q}}

\newcommand{\lambdamp}{\lambda}

\newcommand{\param}{\boldsymbol{\theta}}
\newcommand{\cost}{\mathcal{J}}
\newcommand{\loss}{\mathcal{L}}

\newcommand{\ix}[1]{\mathbf{x}^{(#1)}}
\newcommand{\itarget}[1]{\mathbf{t}^{(#1)}}
\newcommand{\oparam}{\param^{\star}}

\newcommand{\parhead}[1]{\noindent\textbf{#1}\;}

\title{\vspace{-12mm}If Influence Functions are the Answer, Then What is the Question?}

\newcommand*\samethanks[1][\value{footnote}]{\footnotemark[#1]}

\author{Juhan Bae\thanks{Correspondence to: \texttt{jbae@cs.toronto.edu}.}~\thanks{University of Toronto and Vector Institute for Artificial Intelligence.} ,\,
Nathan Ng\samethanks[2]~\thanks{Massachusetts Institute of Technology.},\,
Alston Lo\samethanks[2],\,
Marzyeh Ghassemi\samethanks[3],\,
Roger Grosse\samethanks[2]
}

\begin{document}
\etocdepthtag.toc{mtchapter}
\etocsettagdepth{mtchapter}{subsection}
\etocsettagdepth{mtappendix}{none}

\maketitle

\begin{abstract}

Influence functions efficiently estimate the effect of removing a single training data point on a model's learned parameters. While influence estimates align well with leave-one-out retraining for linear models, recent works have shown this alignment is often poor in neural networks. In this work, we investigate the specific factors that cause this discrepancy by decomposing it into five separate terms. We study the contributions of each term on a variety of architectures and datasets and how they vary with factors such as network width and training time. While practical influence function estimates may be a poor match to leave-one-out retraining for nonlinear networks, we show they are often a good approximation to a different object we term the proximal Bregman response function (PBRF). Since the PBRF can still be used to answer many of the questions motivating influence functions, such as identifying influential or mislabeled examples, our results suggest that current algorithms for influence function estimation give more informative results than previous error analyses would suggest.

\end{abstract}
\section{Introduction}
\label{sec:introduction}

The influence function~\citep{hampel1974influence,cook1979influential} is a classic technique from robust statistics that estimates the effect of deleting a single data example (or a group of data examples) from a training dataset. 
Formally, given a neural network with learned parameters $\oparam$ trained on a dataset $\mathcal{D}$, we are interested in the parameters $\oparam_{-\point}$ learned by training on a dataset $\mathcal{D} - \{\point\}$ constructed by deleting a single training example $\point$ from $\mathcal{D}$. 
By taking the second-order Taylor approximation to the cost function around $\oparam$, influence functions approximate the parameters $\oparam_{-\point}$ without the computationally prohibitive cost of retraining the model. Since~\citet{koh2017understanding} first deployed influence functions in machine learning, influence functions have been used to solve various tasks such as explaining model's predictions~\citep{koh2017understanding,xiaochuang2020explaining}, relabelling harmful training examples~\citep{kong2021resolving}, carrying out data poisoning attacks~\citep{koh2022poisoning}, increasing fairness in models' predictions~\citep{brunet2019understanding,schulam2019trust}, and learning data augmentation techniques~\citep{lee2020learning}.

When the training objective is strongly convex (e.g.,~as in logistic regression with $L_2$ regularization), influence functions are expected to align well with leave-one-out (LOO) or leave-$k$-out retraining~\citep{koh2017understanding,koh2019accuracy,izzo2021approximate}. However,~\citet{basu2020influence} showed that influence functions in neural networks often do not accurately predict the effect of retraining the model and concluded that influence estimates are often ``fragile'' and ``erroneous''. 
Because of the poor match between influence estimates and LOO retraining, influence function methods are often evaluated with alternative metrics such as the recovery rate of maliciously corrupted examples using influence scores~\citep{khanna2019interpreting,koh2017understanding,schioppa2021scaling,karthikeyan2021revisting}. However, these indirect signals make it difficult to develop algorithmic improvements to influence function estimation. If one is interested in improving certain aspects of influence function estimation, such as the linear system solver, it would be preferable to have a well-defined quantity that influence function estimators are approximating so that algorithmic choices could be directly evaluated based on the accuracy of their estimates.

\begin{figure}[t]
    \centering
    \includegraphics[width=1.\textwidth]{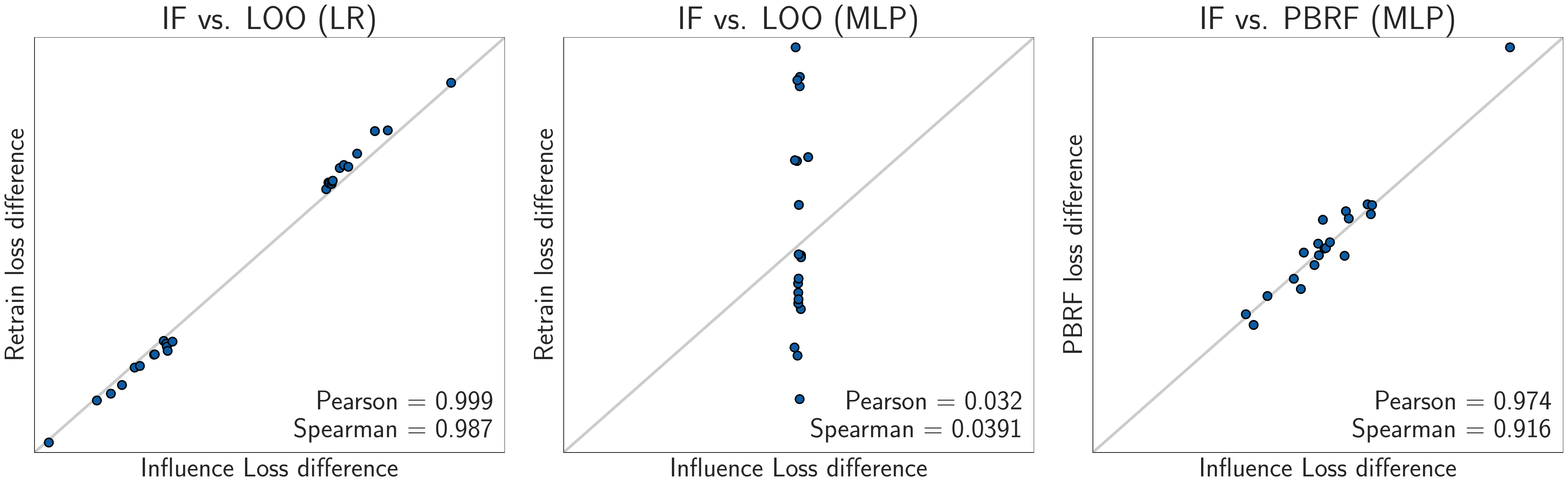}
    \tiny
    \caption{Comparison of test loss differences computed by influence function (IF), leave-one-out (LOO) retraining, and our proximal Bregman response function (PBRF).
    Each point corresponds to the individual effect of removing one training example. 
    Influence estimates align well with true retraining for (\textbf{left}) logistic regression (LR) but poorly for (\textbf{middle}) multilayer perceptrons (MLP). While influence functions in neural networks do not accurately predict the effect of retraining the model, they are still a good approximation to (\textbf{right}) the PBRF.
    }
    \label{fig:corr_lr_mlp}
    \vspace{-0.7cm}
\end{figure}

In this work, we investigate the source of the discrepancy between influence functions and LOO retraining in neural networks. We decompose the discrepancy into five components: (1)~the difference between cold-start and warm-start response functions (a concept elaborated on below), (2)~an implicit proximity regularizer, (3)~influence estimation on non-converged parameters, (4)~linearization, and (5)~approximate solution of a linear system. We empirically evaluate the contributions of each component on binary classification, regression, image reconstruction, image classification, and language modeling tasks and show that, across all tasks, components (1--3) are most responsible for the discrepancy between influence functions and LOO retraining. We further investigate how the contribution of each component changes in response to the change in network width and depth, weight decay, training time, damping, and the number of data points being removed.

Moreover, we show that while influence functions for neural networks are often a poor match to LOO retraining, they are a much better match to what we term the \emph{proximal Bregman response function (PBRF)}. 
Intuitively, the PBRF approximates the effect of removing a data point while trying to keep the predictions consistent with those of the (partially) trained model. From this perspective, we reframe misalignment components (1--3) as simply reflecting the difference between LOO retraining and the PBRF. The gap between the influence function estimate and the PRBF only comes from sources (4) and (5), which we found empirically to be at least an order of magnitude smaller for most neural networks. As a result, on a wide variety of tasks, we show that influence functions closely align with the PBRF while failing to approximate the effect of retraining the model, as shown in Figure~\ref{fig:corr_lr_mlp}. The PBRF can be used for many of the same use cases that have motivated influence functions, such as finding influential or mislabeled examples and carrying out data poisoning attacks~\citep{koh2017understanding,koh2022poisoning}, and can therefore be considered an alternative
to LOO retraining as a gold standard for evaluating influence functions. Hence, we conclude that influence functions applied to neural networks are not inherently ``fragile'' as is often believed~\citep{basu2020influence}, but instead can be seen as giving accurate answers to a different question.

\section{Related Work}

Instance-based interpretability methods are a class of techniques that explain a model's predictions in terms of the examples on which the model was trained.
Methods of this type include \texttt{TracIn}~\citep{pruthi2020estimating}, Representer Point Selection~\citep{yeh2018representer}, Grad-Cos and Grad-Dot~\citep{charpiat2019similarity, hanawa2021evaluation}, \texttt{MMD-critic}~\citep{kim2016examples}, unconditional counterfactual explanations~\citep{wachter2018counterfactual}, and of central focus in this paper, influence functions. 
Since its adoption in machine learning by~\citet{koh2017understanding}, multiple extensions and improvements upon influence functions have also been proposed, such as variants that use Fisher kernels~\citep{khanna2019interpreting}, higher-order approximations~\citep{basu2020second},
tricks for faster and scalable inference~\citep{guo2021fastif, schioppa2021scaling}, group influence formulations~\citep{koh2019accuracy,basu2020second}, and relative local weighting~\citep{barshan2020relatif}.
However, many of these methods rely on the same strong assumptions made in the original influence function derivation that the objective needs to be strongly convex and influence functions must be computed on the optimal parameters. 

In general, influence functions are assumed to approximate the effects of leave-one-out (LOO) retraining from scratch, the parameters of the network that are trained without a data point of interest. Hence, measuring the quality of influence functions is often performed by analyzing the correlation between LOO retraining and influence function estimations~\citep{koh2017understanding,basu2020influence,basu2020second,yang2022understanding}. However, recent empirical analyses have demonstrated the fragility of influence functions and a fundamental misalignment between their assumed and actual effects~\citep{basu2020influence,ghorbani2019interpretation,karthikeyan2021revisting}. For example,~\citet{basu2020influence} argued that the accuracy of influence functions in deep networks is highly sensitive to network width and depth, weight decay strength, inverse-Hessian vector product estimation methodology, and test query point by measuring the alignment between influence functions and LOO retraining. Because of the inherent misalignment between influence estimations and LOO retraining in neural networks, many works often evaluate the accuracy of the influence functions on an alternative metric, such as the recovery rate of maliciously mislabelled or poisoned data using influence functions~\citep{khanna2019interpreting,koh2017understanding,schioppa2021scaling,karthikeyan2021revisting}. In this work, instead of interpreting the misalignment between influence functions and LOO retraining as a failure, we claim that it simply reflects that influence functions answer a different question than is typically assumed.

\section{Background}
\label{sec:background}

Consider a prediction task from an input space $\is$ to a target space $\ts$ where we are given a finite training dataset $\trainset = \{(\inp^{(i)}, \target^{(i)})\}_{i=1}^N$. Given a data point $\point = (\inp, \target)$, let $\out = f(\param, \inp)$ be the prediction of the network parameterized by $\param \in \mathbb{R}^d$ and $\loss(\out, \target)$ be the loss (e.g.,~squared error or cross-entropy). We aim to solve the following optimization problem:
\begin{align}
    \oparam = \argmin_{\param \in \mathbb{R}^d} \cost(\param) = \argmin_{\param \in \mathbb{R}^d} \frac{1}{N} \sum_{i=1}^N \mathcal{L} (f(\param, \ix{i}), \itarget{i}),
    \label{eq:cost_reg}
\end{align}
where $\cost(\cdot)$ is the cost function. If the regularization (e.g.,~$L_2$ regularization) is imposed in the cost function, we fold the regularization terms into the loss function. We summarize the notation used in this paper in Appendix~\ref{app:notation}. 

\subsection{Downweighting a Training Example}
\label{subsec:downweighting}

The training objective in Eqn.~\ref{eq:cost_reg} aims to find the parameters that minimize the average loss on all training examples.
Herein, we are interested in studying the change in optimal model parameters when a particular training example $\point = (\inp, \target) \in \trainset$ is removed from the training dataset, or more generally, when the data point $\point$ is downweighted by an amount $\epsilon \in \mathbb{R}$. 
Formally, this corresponds to minimizing the following downweighted objective:
\begin{align}
    \oparam_{-\point, \epsilon} = \argmin_{\param \in \mathbb{R}^d} \ucost_{-\point}(\param, \epsilon) = \argmin_{\param \in \mathbb{R}^d} \cost(\param) - \loss(f(\param, \inp), \target) \epsilon.
    \label{eq:downweight_objective}
\end{align}
When $\epsilon = \sfrac{1}{N}$, the downweighted objective reduces to the cost over the dataset with the example $\point$ removed, up to a constant factor. 
To see how the optimum of the downweighted objective responds to changes in the downweighting factor $\epsilon$, we define the \emph{response function} $r^{\star}_{-\point}\colon \mathbb{R} \to \mathbb{R}^d$ by:
\begin{align}
    r^{\star}_{-\point} (\epsilon) = \argmin_{\param \in \mathbb{R}^d} \ucost_{-\point}(\param, \epsilon),
\end{align}
where we assume that the downweighted objective is strongly convex and hence the solution to the downweighted objective is unique given some factor $\epsilon$.
Under these assumptions, note that $r^{\star}_{-\point} (0) = \oparam$ and the response function is differentiable at $0$ by the Implicit Function Theorem~\citep{krantz2002implicit,griewank2008evaluating}.
Influence functions defined as approximate the response function by performing a first-order Taylor expansion around $\epsilon_0 = 0$:
\begin{align}
    r_{-\point, \text{lin}}^{\star} (\epsilon) &= r^{\star}_{-\point} (\epsilon_0) +  \frac{\mathrm{d} r^{\star}_{-\point}}{\mathrm{d} \epsilon} \biggr\rvert_{\epsilon = \epsilon_0} (\epsilon - \epsilon_0) = \oparam + (\nabla^2_{\param} \cost(\oparam))^{-1} \nabla_{\param} \loss(f(\oparam, \inp), \target) \epsilon.
    \label{eq:first_order_response}
\end{align}
We refers readers to \citet{van2000asymptotic} and Appendix~\ref{app:inf-derivation} for a detailed derivation. 
The optimal parameters trained without $\point$  
can then be approximated by plugging in $\epsilon = \sfrac{1}{N}$ to Eqn.~\ref{eq:first_order_response}. 
Influence functions can further approximate the loss of a particular test point $\point_{\text{test}} = (\inp_{\text{test}}, \target_{\text{test}})$ when a data point $\point$ is eliminated from the training set using the chain rule~\citep{koh2017understanding}:
\begin{align}
    \begin{split}
    &\loss(f(r^{\star}_{-\point, \text{lin}} \left( \sfrac{1}{N} \right), \inp_{\text{test}}), \target_{\text{test}}) \\
    &\quad \approx \loss(f(\oparam, \inp_{\text{test}}), \target_{\text{test}}) + \frac{1}{N} \nabla_{\param} \loss(f(\oparam, \inp_{\text{test}}), \target_{\text{test}})^{\top}  \frac{\mathrm{d} r^{\star}_{\point}}{\mathrm{d} \epsilon} \biggr\rvert_{\epsilon = 0}  \\
    &\quad = \loss(f(\oparam, \inp_{\text{test}}), \target_{\text{test}}) + \frac{1}{N} \nabla_{\param} \loss(f(\oparam, \inp_{\text{test}}), \target_{\text{test}})^{\top} (\nabla^2_{\param} \cost(\oparam))^{-1} \nabla_{\param} \loss(f(\oparam, \inp), \target).
    \end{split}
    \label{eq:influence_test_loss}
\end{align}

\subsection{Influence Function Estimation in Neural Networks}
\label{subsec:influence_nn}
Influence functions face two main challenges when deployed on neural networks. First, the influence estimation (shown in Eqn.~\ref{eq:first_order_response}) requires computing an inverse Hessian-vector product (\texttt{iHVP}). Unfortunately, storing and inverting the Hessian requires $O(d^3)$ operations and is infeasible to compute for modern neural networks. Instead,~\citet{koh2017understanding} tractably approximate the \texttt{iHVP} using truncated non-linear conjugate gradient (\texttt{CG})~\citep{martens2010deep} or the \texttt{LiSSA} algorithm~\citep{agarwal2016second}. Both approaches avoid explicit computation of the Hessian inverse (see Appendix~\ref{app:efficient_computation} for details) and only require $O(Nd)$ operations to approximate the influence function.

Second, the derivation of influence functions assumes a strongly convex objective, which is often not satisfied for neural networks. The Hessian may be singular, especially when the parameters have not fully converged, due to non-positive eigenvalues. 
To enforce positive-definiteness of the Hessian,~\citet{koh2017understanding} add a damping term in the \texttt{iHVP}.~\citet{teso2021interactive} further approximate the Hessian with the Fisher information matrix (which is equivalent to the Gauss-Newton Hessian~\citep{martens2014new} for commonly used loss functions such as cross-entropy) as follows:
\begin{align} 
    r^{\star}_{-\point, \text{damp}, \text{lin}} (\epsilon) \approx \oparam + (
    \mathbf{J}_{\mathbf{y}\oparam}^{\top} \mathbf{H}_{\mathbf{y}^{\star}} \mathbf{J}_{\mathbf{y}\oparam}
    + \lambda \mathbf{I})^{-1} \nabla_{\param} \loss(f(\oparam, \inp), \target) \epsilon,
    \label{eq:inf_gn_approx}
\end{align}
where $\mathbf{J}_{\mathbf{y}\oparam}$ is the parameter-output Jacobian and $\mathbf{H}_{\mathbf{y}^{\star}}$ is the Hessian of the cost with respect to the network outputs both evaluated on the optimal parameters $\oparam$.
Here, $\mathbf{G}^{\star} = \mathbf{J}_{\mathbf{y}\oparam}^{\top} \mathbf{H}_{\mathbf{y}^{\star}} \mathbf{J}_{\mathbf{y}\oparam}$ is the Gauss-Newton Hessian (GNH) and $\lambda > 0$ is a damping term to ensure the invertibility of GNH. 
Unlike the Hessian, the GNH is guaranteed to be positive semidefinite as long as the loss function is convex as a function of the network outputs~\citep{martens2010deep}.
\section{Understanding the Discrepancy between Influence Function and LOO Retraining in Neural Networks}
\label{sec:source_of_error}

In this section, we investigate several factors responsible for the misalignment between influence functions and LOO retraining. Specifically, we decompose the misalignment into five separate terms: (1) the warm-start gap, (2) proximity gap, (3) non-convergence gap, (4) linearization error, and (5) solver error. This decomposition captures all approximations and assumption violations when deploying influence functions in neural networks. By summing the parameter (or outputs) differences introduced by each term we can bound the parameter (or outputs) difference between LOO retraining and influence estimates.
We use the term ``gap'' rather than ``error'' for the first three terms to emphasize that they reflect differences between solutions to different influence-related questions, rather than actual errors. For all models we investigate, we find that the first three sources dominate the misalignment, indicating that the misalignment reflects not algorithmic errors but rather the fact that influence function estimators are answering a different question from what is normally assumed. 
All proximal objectives are summarized in Table~\ref{tab:response_list} and we provide the derivations in Appendix~\ref{app:inf-derivation}.

\subsection{Warm-Start Gap: Non-Strongly Convex Training Objective}

\begin{wrapfigure}[22]{R}{0.35\textwidth}
    \centering
    \vspace{-0.9cm}
    \includegraphics[width=0.35\textwidth]{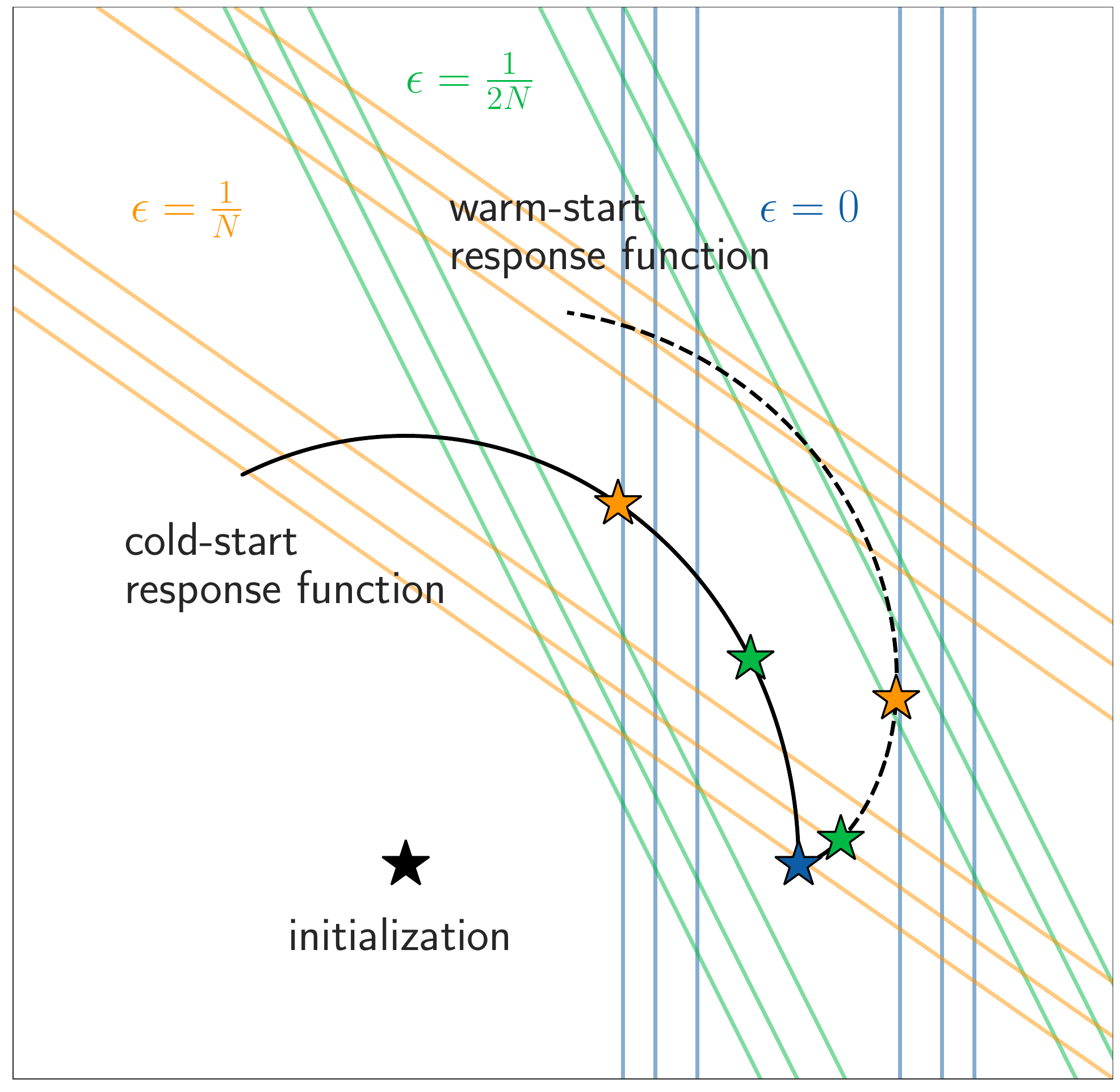}
    \vspace{-0.46cm}
    \small
    \caption{Cold-start (initialized from black star) and warm-start (initialized from blue star) response functions for quadratic cost function. Each contour represents the cost function at some $\epsilon$. Because gradient descent converges to a minimum-norm solution, the warm-start and cold-start optima are not equivalent.
    }
    \label{fig:cold-warm-viz}
\end{wrapfigure}

By taking a first-order Taylor approximation of the response function at $\epsilon_0 = 0$ (Eqn.~\ref{eq:first_order_response}), influence functions approximate the effect of removing a data point $\point$ at a local neighborhood of the optimum $\oparam$. Hence, influence approximation has a more natural connection to the retraining scheme that initializes the network at the current optimum $\oparam$ (\emph{warm-start retraining}) than the scheme that initializes the network randomly (\emph{cold-start retraining}). The warm-start optimum is equivalent to the cold-start optimum when the objective is strongly convex (where the solution to the response function is unique), making the influence estimation close to the LOO retraining on logistic regression with $L_2$ regularization.

However, the equivalence between warm-start and cold-start optima is not typically guaranteed in neural networks~\citep{ash2020warm,vicol2021implicit}. Particularly, in the overparametrized regime ($N < d$), neural networks exhibit multiple global optima, and their converged solutions depend highly on the specifics of the optimization dynamics~\citep{lee2019wide,arora2019fine,bartlett2020benign,amari2020does}. For quadratic cost functions, gradient descent with initialization $\param^0$ converges to the optimum that achieves the minimum $L_2$ distance from $\param^0$~\citep{hastie2022surprises}.
This phenomenon of the converged parameters being dependent on the initialization hinders influence functions from accurately predicting the effect of retraining the model from scratch as shown in Figure~\ref{fig:cold-warm-viz}. We denote the discrepancy between cold-start and warm-start optima as \textbf{warm-start gap}.

\subsection{Proximity Gap: Addition of Damping Term in \texttt{iHVP}}

In practical settings, we often impose a damping term (Eqn.~\ref{eq:inf_gn_approx}) in influence approximations to ensure that the Hessian of the cost is positive-definite and hence invertible. As adding a damping term in influence estimations is equivalent to adding $L_2$ regularization to the cost function~\citep{martens2010deep}, when damping is used, influence functions can be seen as linearizing the following \emph{proximal response function} at $\epsilon_0 = 0$:
\begin{align}
    r_{-\point, \text{damp}} ^{\star}(\epsilon) = \argmin_{\param \in \mathbb{R}^d} \ucost_{-\point} (\param, \epsilon) + \frac{\lambdamp}{2}\|\param - \oparam\|^2.
    \label{eq:damped_warm_optimum}
\end{align}
Note that our use of ``proximal'' is based on the notion of proximal equilibria~\citep{farnia2020gans}. Intuitively, the proximal objective in Eqn.~\ref{eq:damped_warm_optimum} not only minimizes the downweighted objective but also encourages the parameters to stay close to the optimal parameters at $\epsilon_0 = 0$. Hence, when the damping term is used in the \texttt{iHVP}, influence functions aim at approximating the warm-start retraining scheme with a proximity term that penalizes the $L_2$ distance between the new estimate and the optimal parameters. We call the discrepancy between the warm-start and proximal warm-start optima the \textbf{proximity gap}.

Interestingly, past works have observed that for quadratic cost functions, early stopping has a similar effect to $L_2$ regularization~\citep{vicol2021implicit, ali2019continuous}. Therefore, the proximal response function can be thought of as capturing how gradient descent will respond to a dataset perturbation if it takes only a limited number of steps starting from the warm-start solution.

\subsection{Non-Convergence Gap: Influence Estimation on Non-Converged Parameters}
\label{sec:non_converge}

\begin{wrapfigure}[16]{R}{0.45\textwidth}
    \centering
    \vspace{-1cm}
    \includegraphics[width=0.45\textwidth]{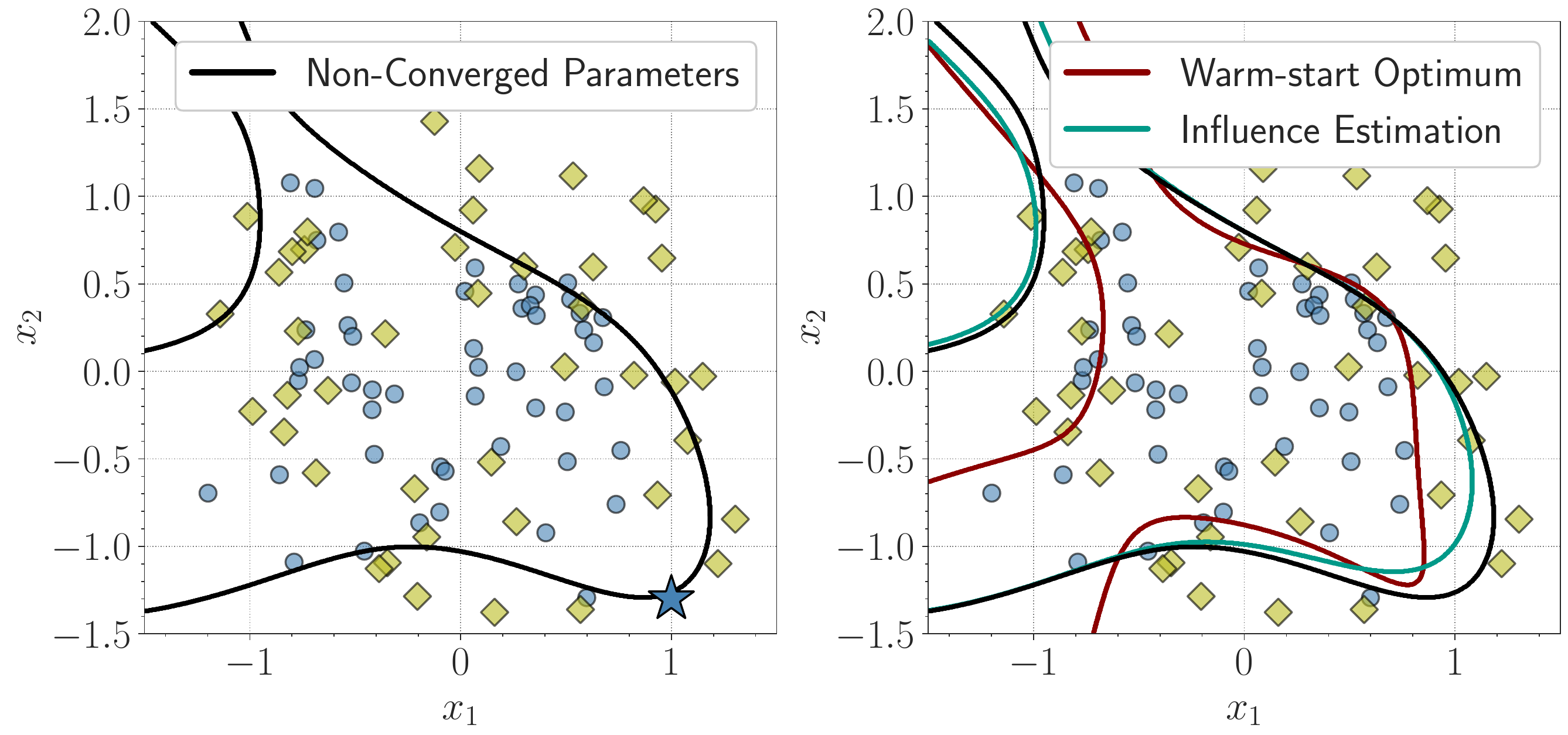}
    \vspace{-0.5cm}
    \small
    \caption{Decision boundaries for a partially trained binary classifier. We consider removing a data point located at right-bottom corner denoted as {\color{dkblue} $\star$}. While the influence estimation makes a local change on the data point of interest, the LOO retraining globally updates the parameters to better fit other data points (a nuisance from the perspective of understanding influence).}
    \label{fig:decision_boundary}
\end{wrapfigure}

Thus far, our analysis has assumed that influence functions are computed on fully converged parameters $\oparam$ at which the gradient of the cost is \textbf{0}. However, in neural network training, we often terminate the optimization procedure before reaching the exact optimum due to several reasons, including having limited computational resources or to avoid overfitting~\citep{bengio2012practical}. In such situations, much of the change in the parameters from LOO retraining simply reflects the effect of training for longer, rather than the effect of removing a training example, as illustrated in Figure~\ref{fig:decision_boundary}. What we desire from influence functions is to understand the effect of removing the training example; the effect of extended training is simply a nuisance. Therefore, to the extent that this factor contributes to the misalignment between influence functions and LOO retraining, influence functions are arguably \emph{more useful} than LOO retraining.

Since training the network to convergence may be impractical or undesirable, we instead modify the response function by replacing the original training objective with a similar one \emph{for which the (possibly non-converged) final parameters $\param^s$ are optimal}. Here, we assume the loss function is convex as a function of the network outputs; this is true for commonly used loss functions such as squared error or cross-entropy. We replace the training loss with a term that penalizes mismatch to the predictions made by $\param^s$ (hence implying that $\param^s$ is optimal). Our \emph{proximal Bregman response function (PBRF)} is defined as follows:
\begin{align}
    r^b_{-\point, \text{damp}} (\epsilon) = \argmin_{\param \in \mathbb{R}^d} \frac{1}{N} \sum_{i=1}^N D_{\loss^{(i)}} (f(\param, \inp^{(i)}), f(\param^s, \inp^{(i)})) - \loss(f(\param, \inp), \target) \epsilon 
    + \frac{\lambdamp}{2} \| \param - \param^s \|^2,
    \label{eq:pbrf}
\end{align}
where $D_{\loss^{(i)}} (\cdot, \cdot)$ is the Bregman divergence defined as:
\begin{align}
    D_{\loss^{(i)}} (\mathbf{y}, \mathbf{y}^s) = \loss(\mathbf{y}, \mathbf{t}^{(i)}) - \loss(\mathbf{y}^s, \mathbf{t}^{(i)}) -  \nabla_{\mathbf{y}} \loss(\mathbf{y}^s, \mathbf{t}^{(i)})^\top (\mathbf{y} - \mathbf{y}^s ).
\end{align}
The PBRF defined in Eqn.~\ref{eq:pbrf} is composed of three terms. The first term measures the functional discrepancy between the current estimate and the parameters $\param^s$ in Bregman divergence, and its role is to prevent the new estimate from drastically altering the predictions on the training dataset. One way of understanding this term in the cases of squared error or cross-entropy losses is that it is equivalent to the training error on a dataset where the original training labels are replaced with soft targets obtained from the predictions made by $\param^s$. The second term is the negative loss on the data point $\point = (\inp, \target)$, which aims to respond to the deletion of a training example. The final term is simply the proximity term described before. In Appendix~\ref{app:suboptimal_response_function}, we further show that the influence function on non-converged parameters is equivalent to the first-order approximation of PBRF instead of the first-order approximation of proximal response function for linear models.

Rather than computing the LOO retrained parameters by performing $K$ additional optimization steps under the original training objective, we can instead perform $K$ optimization steps under the proximal Bregman objective. The difference in the proximal response function and the PBRF is what we call the \textbf{non-convergence gap}.

\subsection{Linearization Error: A First-order Taylor Approximation of the Response Function.}
The key idea behind influence functions is the linearization of the response function. Recall that influence functions with the GNH approximation and a damping term on possibly non-converged parameters are defined as:
\begin{align} 
    r^{\star}_{-\point, \text{damp}, \text{lin}} (\epsilon) \approx \param^s + (
    \mathbf{J}_{\mathbf{y}\param^s}^{\top} \mathbf{H}_{\mathbf{y}^{s}} \mathbf{J}_{\mathbf{y}\param^s}
    + \lambda \mathbf{I})^{-1} \nabla_{\param} \loss(f(\param^s, \inp), \target) \epsilon.
    \label{eq:no_gnh}
\end{align}
As opposed to Eqn.~\ref{eq:inf_gn_approx}, influence functions are computed on a possibly non-converged parameters $\param^s$ instead of the optimal parameters $\oparam$. To simulate the local approximations made in influence functions, we define the linearized PBRF as:
\begin{align}
\begin{split}
    r^b_{-\point, \text{damp}, \text{lin}} (\epsilon) = \argmin_{\param \in \mathbb{R}^d} \frac{1}{N} &\sum_{i=1}^N D_{\loss^{(i)}_{\text{quad}}} (f_{\text{lin}}(\param, \inp^{(i)}), f(\param^s, \inp^{(i)})) \\
    &\quad- \nabla_{\param}\loss(f(\param^s, \inp), \target)^{\top} \param \epsilon + \frac{\lambdamp}{2} \| \param - \param^s \|^2,
    \label{eq:lin_pbrf}
\end{split}
\end{align}
where $\loss_{\text{quad}} (\cdot, \cdot)$ is the second-order expansion of the loss around $\mathbf{y}^s$ and $f_{\text{lin}}(\cdot, \cdot)$ is the linearization of the network outputs with respect to the parameters as shown below:
\begin{align}
    \loss_{\text{quad}} (\mathbf{y}, \mathbf{t})= \loss (\mathbf{y}^s, \mathbf{t}) + \nabla_{\mathbf{y}}& \loss(\mathbf{y}^s, \mathbf{t})^{\top} (\mathbf{y} - \mathbf{y}^s) + (\mathbf{y} - \mathbf{y}^s)^{\top} \mathbf{H}_{\mathbf{y}^{s}} (\mathbf{y} - \mathbf{y}^s).\\
    f_{\text{lin}}(\param, \inp^{(i)}) &= f(\param^s, \inp^{(i)}) + \mathbf{J}_{\mathbf{y}^{(i)} \param^s} (\param - \param^s).
\end{align}
The optimal solution to the linearized PBRF is equivalent to the influence estimation at the parameters $\param^s$ with the GNH approximation and a damping term $\lambda$ (see Appendix~\ref{app:lin_response_function_gnh} for the derivation).
As the linearized PBRF relies on several local approximations, the linearization error increases when the downweighting factor magnitude $|\epsilon|$ is large or the PBRF is highly non-linear. We refer to the discrepancy between the PBRF and linearized PBRF as the \textbf{linearization error}.

\begin{table}[t]
    \centering
    \resizebox{\columnwidth}{!}{
    \begin{tabular}{lcc}
    \toprule
    \textbf{Error} & \textbf{Objective} &  \textbf{Init}\\ \midrule
    Cold-start    & $\cost(\param) - \loss(f(\param, \inp), \target) \epsilon$  & $\param^0$\\[0.25em]
    + Warm-start & $\cost(\param) - \loss(f(\param, \inp), \target) \epsilon$  & $\param^s$ \\[0.25em]
    + Proximity  & $ \cost(\param) - \loss(f(\param, \inp), \target) \epsilon+ \frac{\lambdamp}{2} \|\param - \param^s\|^2$ & $\param^s$  \\[0.25em]
    + Non-Convergence    & $ \frac{1}{N} \sum_{i=1}^N D_{\loss^{(i)}} (f(\param, \inp^{(i)}), f(\param^s, \inp^{(i)})) - \loss(f(\param, \inp), \target) \epsilon + \frac{\lambdamp}{2} \| \param - \param^s \|^2$  & $\param^s$ \\[0.25em]
    + Linearization   & $ \frac{1}{N} \sum_{i=1}^N D_{\loss^{(i)}_{\text{quad}}} (f_{\text{lin}}(\param, \inp^{(i)}), f(\param^s, \inp^{(i)})) - \nabla_{\param} \loss(f(\param^s, \inp), \target)^{\top} \param \epsilon + \frac{\lambdamp}{2} \| \param - \param^s \|^2$  & $\param^s$ \\
    \bottomrule
    \end{tabular}}
    \vspace{-0.1cm}
    \caption{Summary of proximal objectives that influence functions aim to approximate when the network is non-strongly convex, a damping term is used, and influence functions are computed on non-converged parameters.
    }
    \label{tab:response_list}
    \vspace{-1.5\baselineskip}
\end{table}

\subsection{Solver Error: A Crude Approximation of \texttt{iHVP}}
As the precise computation of the \texttt{iHVP} is computationally infeasible, in practice, we use truncated \texttt{CG} or \texttt{LiSSA} to efficiently approximate influence functions \citep{koh2017understanding}. Unfortunately, these efficient linear solvers introduce additional error by crudely approximating the \texttt{iHVP}. 
Moreover, different linear solvers can introduce specific biases in the influence estimation. For example, \citet{vicol2022implicit} show that the truncated \texttt{LiSSA} algorithm implicitly adds an additional damping term in the \texttt{iHVP}. 
We use \textbf{solver error} to refer to the difference between the linearized PBRF and the influence estimation computed by a linear solver. 
\section{PBRF: The Question Influence Functions are Really Answering}
\label{sec:decompose}

The PBRF (Eqn.~\ref{eq:pbrf}) approximates the effect of removing a data point while trying to keep the predictions consistent with those of the (partially) trained model.
Since the discrepancy between the PBRF and influence function estimates is only due to the linearization and solver errors, the PBRF can be thought of as better representing the question that influence functions are trying to answer. 

Reframing influence functions in this way means that the PBRF can be regarded as a gold-standard ground truth for evaluating influence function approximation.
Existing analyses of influence functions~\citep{basu2020influence} 
rely on generating LOO retraining ground truth estimates by 
imposing strong $L_2$ regularization or training till convergence without early stopping to implicitly reduce warm-start and non-convergence gaps.
However, these conditions do not accurately reflect the typical way neural networks are trained in practice.
In contrast, our PBRF formulation 
does not require the addition of any regularizers or modified training regimes and can be easily optimized.

In addition, although the PBRF may not necessarily align with LOO retraining due to the warm-start, proximity, and non-convergence gaps,
the motivating use cases for influence functions typically do not rely on exact LOO retraining. 
This means that the PBRF can be used in place of LOO retraining for many tasks such as identifying influential or mislabelled examples (as shown in Appendix~\ref{app:mislabelled}). In these cases, influence functions are still useful since they provide an efficient way of approximating PBRF estimates.
\section{Experiments}
\label{sec:experiments}

Our experiments investigate the following questions: (1) What factors discussed in Section~\ref{sec:source_of_error} contribute most to the misalignment between influence functions and LOO retraining? (2) While influence functions fail to approximate the effect of retraining, do they accurately approximate the PBRF? (3) How do changes in weight decay, damping, the number of total epochs, and the number of removed training examples affect each source of misalignment?

In all experiments, we first train the base network with the entire dataset to obtain the parameters $\param^s$. We repeat the training procedure 20 times with a different random training example deleted. The cold-start retraining trains the parameters from the same initialization used to train $\param^s$. All proximal objectives are trained with initialization $\param^s$ for 50\% of the epochs used to train the base network.
Lastly, we use the \texttt{LiSSA} algorithm with the GNH approximation to compute influence functions.

Since we are primarily interested in the effect of deleting a data point on model's predictions, we measure the discrepancy of each gap and error using the average $L_2$ distance between networks' outputs $\mathbb{E}_{(\inp, \cdot) \sim \mathcal{D}_{\text{train}}} [ \|f(\param, \mathbf{x}) - f(\param', \mathbf{x})\|]$ on the training dataset. We provide the full experimental set-up and additional experiments
in Appendix~\ref{app:experiment_details} and~\ref{app:additional_results}, respectively.

\subsection{Influence Misalignment Decomposition}

\begin{figure}[t]
    \centering
    \includegraphics[width=0.9\textwidth]{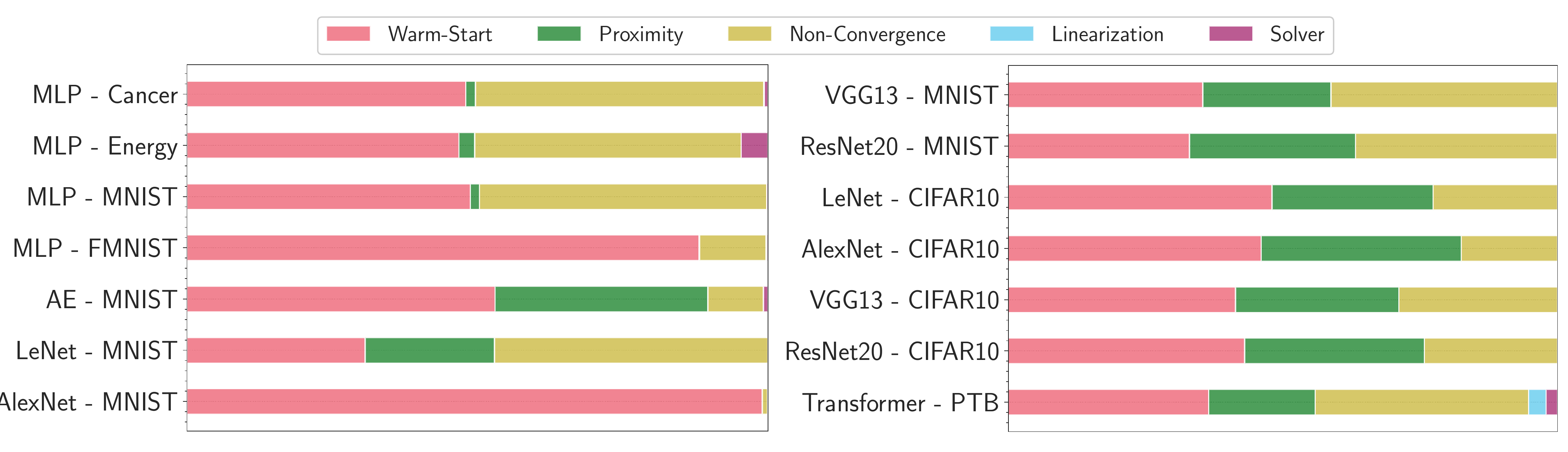}
    \small
    \caption{
    Decomposition of the discrepancy between influence functions and LOO retraining into (1) warm-start gap, (2) proximal gap, (3) non-convergence gap, (4) linearization error, and (5) solver error for each model and dataset. The size of each component is measured by the $L_2$ distance between the networks' outputs on the training dataset.}
    \label{fig:error_decompse}
\end{figure}

We first applied our decomposition to various models trained on a broad range of tasks covering binary classification, regression, image reconstruction, image classification, and language modeling. The summary of our results is provided in Figure~\ref{fig:error_decompse} and Table~\ref{tab:results2} (Appendix~\ref{app:decomp}). Across all tasks, we found that the first three sources (warm-start, proximity, and non-convergence gaps) dominate the misalignment, indicating the discrepancy between influence functions and LOO retraining is mainly caused by influence function estimators answering a different question from what is normally assumed. Small linearization and solver errors indicate that influence functions accurately answers the modified question (PBRF).

\parhead{Logistic Regression.} We analyzed the logistic regression (LR) model trained on the Cancer and Diabetes classification datasets from the UCI collection~\citep{Dua2019}. We trained the model using \texttt{L-BFGS}~\citep{liu1989limited} with $L_2$ regularization of 0.01 and damping term of $\lambdamp = 0.001$. As the training objective is strongly convex and the base model parameters were trained till convergence, we observed that each source of misalignment is significantly low. Hence, in case of logistic regression with $L_2$ regularization, influence functions accurately capture the effect of retraining the model without a data point.

\parhead{Multilayer Perceptron.} 
Next, we applied our analysis to the 2-hidden layer Multilayer Perceptron (MLP) with ReLU activations. We conducted the experiments in two settings: (1) regression on the Concrete and Energy datasets from the UCI collection and (2) image classification on 10\% of the MNIST~\citep{deng2012mnist} and FashionMNIST~\citep{xiao2017fashion} datasets, following the set-up from~\citet{koh2017understanding} and~\citet{basu2020influence}. We trained the networks for 1000 epochs using stochastic gradient descent (SGD) with a batch size of 128 and set a damping strength of $\lambdamp = 0.001$.

As opposed to linear models, MLPs violate the assumptions in the influence derivation and we observed an increase in gaps and errors on all five factors. We observed that warm-start, proximity, and the non-convergence gaps contribute more to the misalignment than linearization and solver errors. 
The average network's predictions for PBRF were similar to that computed by the \texttt{LiSSA} algorithm, demonstrating that influence functions are still a good approximation to PBRF. 

\parhead{Autoencoder.}
Next, we applied our framework to an 8-layer autoencoder (AE) on the full MNIST dataset. We followed the experimental set-up from~\citet{martens2015optimizing}, where the encoder and decoder each consist of 4 fully-connected layers with sigmoid activation functions. We trained the network for 1000 epochs using SGD with momentum. We set the batch size to 1024, used $L_2$ regularization of $10^{-5}$ with a damping factor of $\lambdamp = 0.001$. In accordance with the findings from our MLP experiments, the warm-start, proximity, and non-convergence gaps were more significant than the linearization and solver errors, and influence functions accurately predicted the PBRF.

\begin{wraptable}[12]{r}{0.45\textwidth}
    \vspace{-0.9cm}
    \centering
    \footnotesize
    \resizebox{0.45\textwidth}{!}{%
    \begin{tabular}{@{}ccccccc@{}}
    \toprule
    \textbf{Model} & \multicolumn{2}{c}{\textbf{Cold-Start}} & \multicolumn{2}{c}{\textbf{Warm-Start}} & \multicolumn{2}{c}{\textbf{PBRF}} \\ \cmidrule(l){2-7} 
     & P & S & P & S & P & S \\ \midrule
    MLP & -0.55 & 0.01 & 0.22 & 0.35 & \textbf{0.98}  &  \textbf{0.99}\\ 
    LeNet & -0.19 & 0.12 & 0.32  & 0.25  & \textbf{0.93}  & \textbf{0.52}  \\ 
    AlexNet & -0.16 & -0.08 & 0.51 & 0.58 & \textbf{0.99} & \textbf{0.99} \\
    VGG13 & 0.45 & -0.07 & -0.28 & -0.51 & \textbf{0.98} & \textbf{0.77} \\ 
    ResNet-20 & 0.09 & -0.06 & 0.02  & 0.09 & \textbf{0.81}  & \textbf{0.76}\\ \bottomrule
    
    \end{tabular}
    }
    \vspace{-0.2cm}
    \caption{Comparison of test loss differences computed by influence function, cold-start retraining, warm-start retraining, and PBRF on MNIST dataset. We show Pearson (P) and Spearman rank-order (S) correlation when compared to influence estimates.
    }
    \label{table:corr_inf}
\end{wraptable}
\parhead{Convolutional Neural Networks.}
To investigate the source of discrepancy on larger-scale networks, we trained a set of convolutional neural networks of increasing complexity and size. Namely, LeNet~\citep{lecun1998gradient}, AlexNet~\citep{krizhevsky2012imagenet}, VGG13~\cite{simonyan2014very}, and ResNet-20~\citep{he2015deep} were trained on 10\% of the MNIST dataset and the full CIFAR10~\citep{Krizhevsky2009learning} dataset. We trained the base network for 200 epochs on both datasets with a batch size of 128. For MNIST, we kept the learning rate fixed throughout training, while for CIFAR10, we decayed the learning rate by a factor of 5 at epochs 60, 120, and 160, following~\citet{zagoruyko2016wide}. We used $L_2$ regularization with strength $5 \cdot 10^{-4}$ and a damping factor of $\lambdamp =0.001$. Consistent with the findings from our MLP and autoencoder experiments, the first three gaps were more significant than linearization and solver errors.  

We further compared influence functions' approximations on the difference in test loss when a random training data point is removed with the value obtained from cold-start retraining, warm-start retraining, and PBRF in Table~\ref{table:corr_inf}. We used both Pearson~\citep{sedgwick2012pearson} and Spearman rank-order correlation~\citep{spearman1961proof} to measure the alignment. While the test loss predicted by influence functions do not align well with the values obtained by cold-start and warm-start retraining schemes, they show high correlations when compared to the estimates given by PBRF.

\parhead{Transformer.} 
Finally, we trained 2-layer Transformer language models on the Penn Treebank (PTB)~\citep{marcus1993building} dataset. We set the number of hidden dimensions to $256$ and the number of attention heads to $2$. As we observed that model overfits after a few epochs of training, we trained the base network for 10 epochs using Adam. Notably, we observed that the non-convergence gap had the most considerable contribution to the discrepancy between influence functions and LOO retraining. Consistent with our previous findings, the first tree gaps had more impact on the discrepancy compared to linearization and solver errors.

\subsection{Factors in Influence Misalignment}
\label{sec:factors_exp}

\begin{figure}[t]
\centering
\begin{subfigure}[t]{0.32\textwidth}
    \centering
    \includegraphics[height=1.18in]{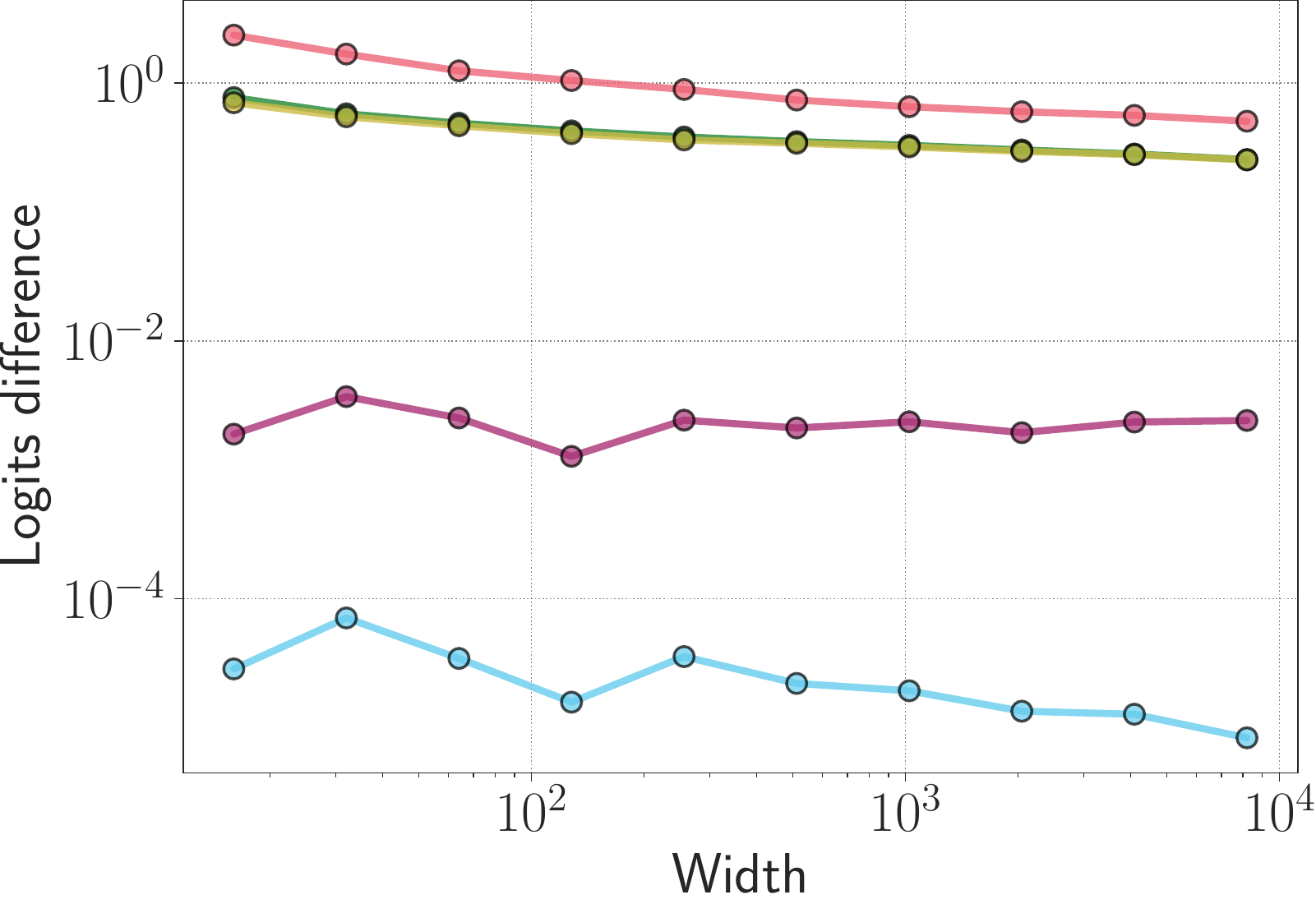}
    \vspace{-0.1cm}
    \caption{{\footnotesize Effect of width}}
    \label{subfig:cnn-mnist}
\end{subfigure}
\begin{subfigure}[t]{0.32\textwidth}
    \centering
    \includegraphics[height=1.18in]{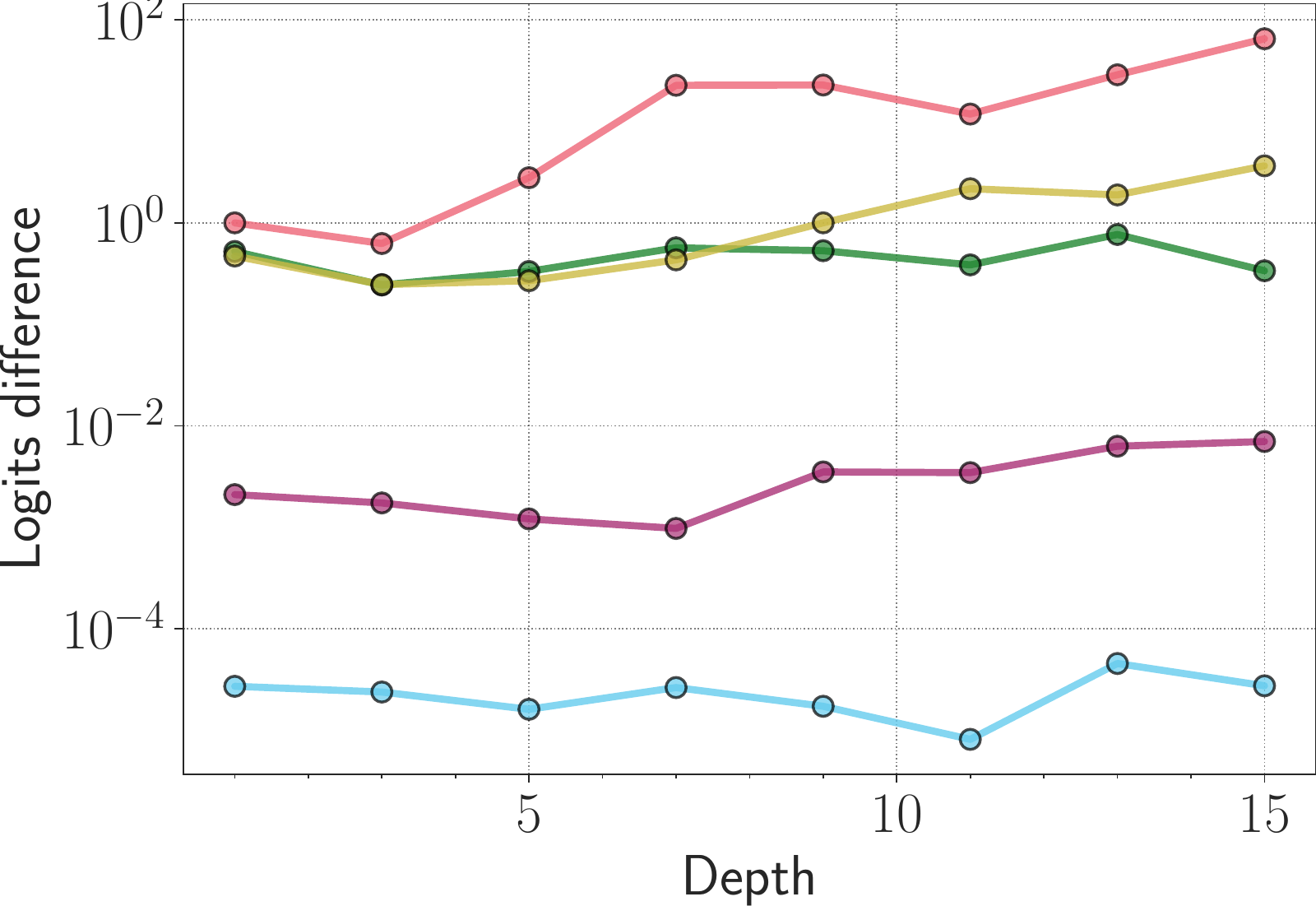}
    \vspace{-0.1cm}
    \caption{{\footnotesize Effect of depth}}
    \label{subfig:cnn-fmnist}
\end{subfigure}
\begin{subfigure}[t]{0.32\textwidth}
    \centering
    \includegraphics[height=1.18in]{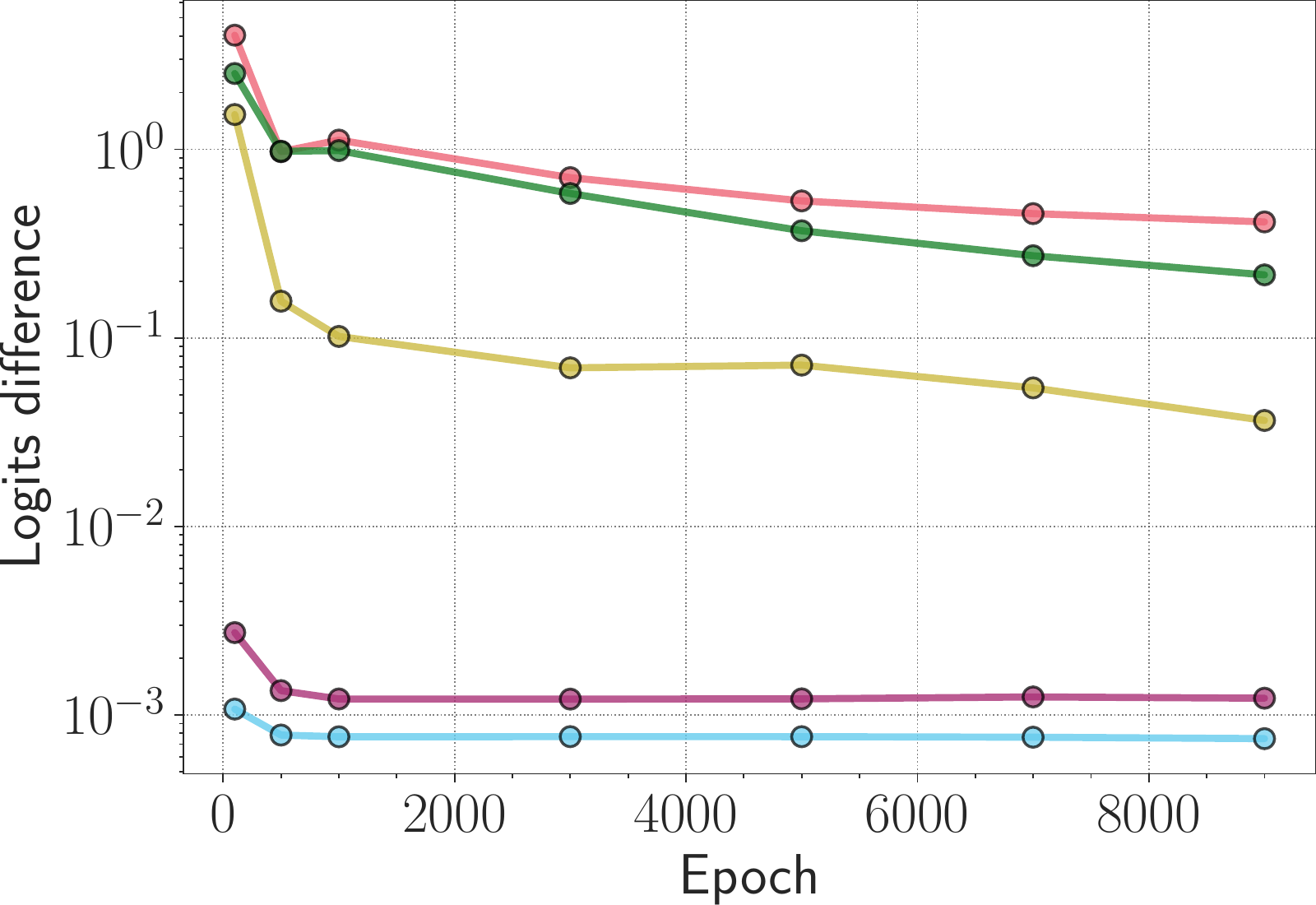}
    \vspace{-0.1cm}
    \caption{{\footnotesize Effect of training time}}
    \label{subfig:resnet8-cifar}
\end{subfigure}
\begin{subfigure}[t]{0.32\textwidth}
    \centering
    \includegraphics[height=1.18in]{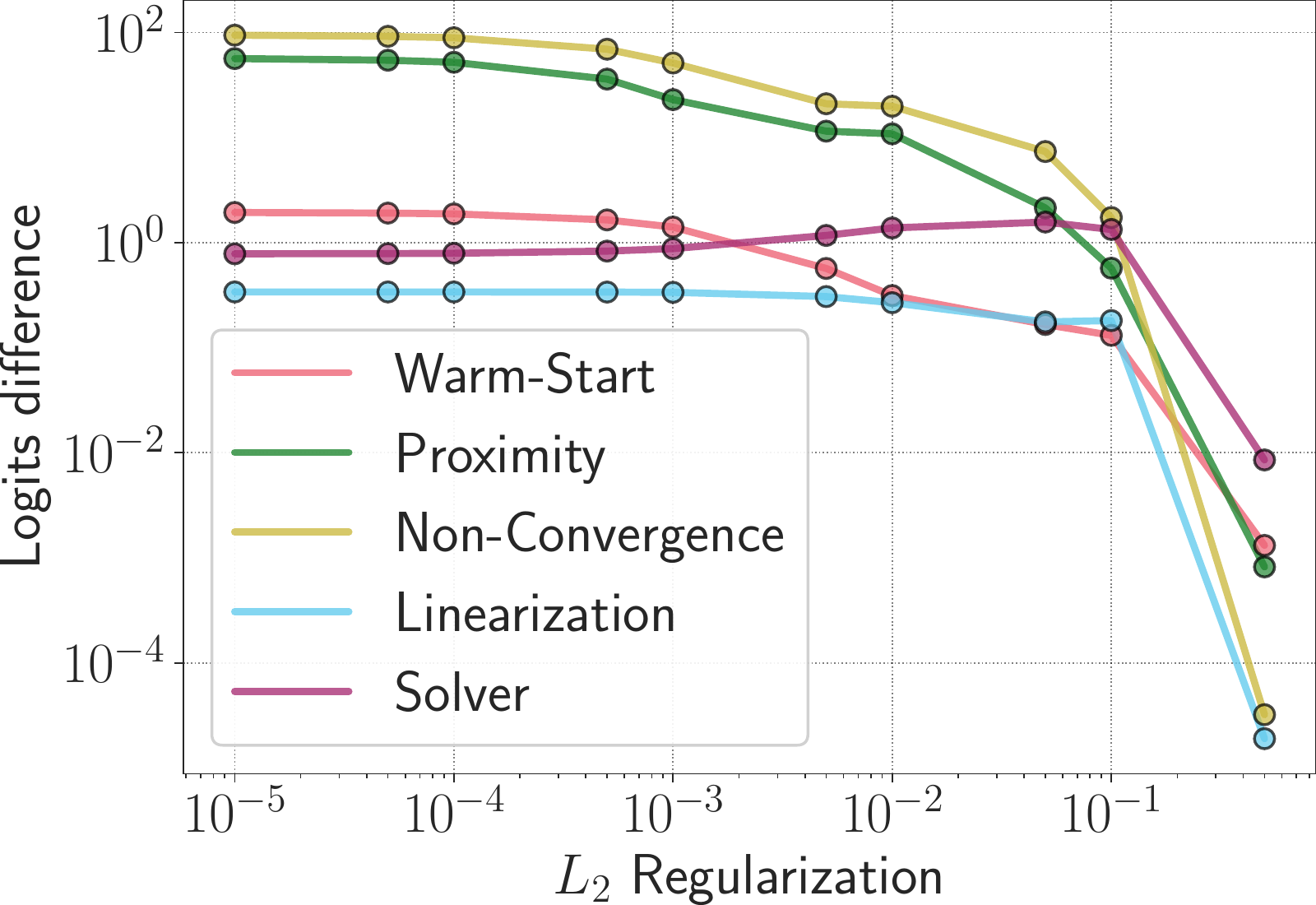}
    \vspace{-0.1cm}
    \caption{{\footnotesize Effect of $L_2$ regularization}}
\end{subfigure}
\begin{subfigure}[t]{0.32\textwidth}
    \centering
    \includegraphics[height=1.18in]{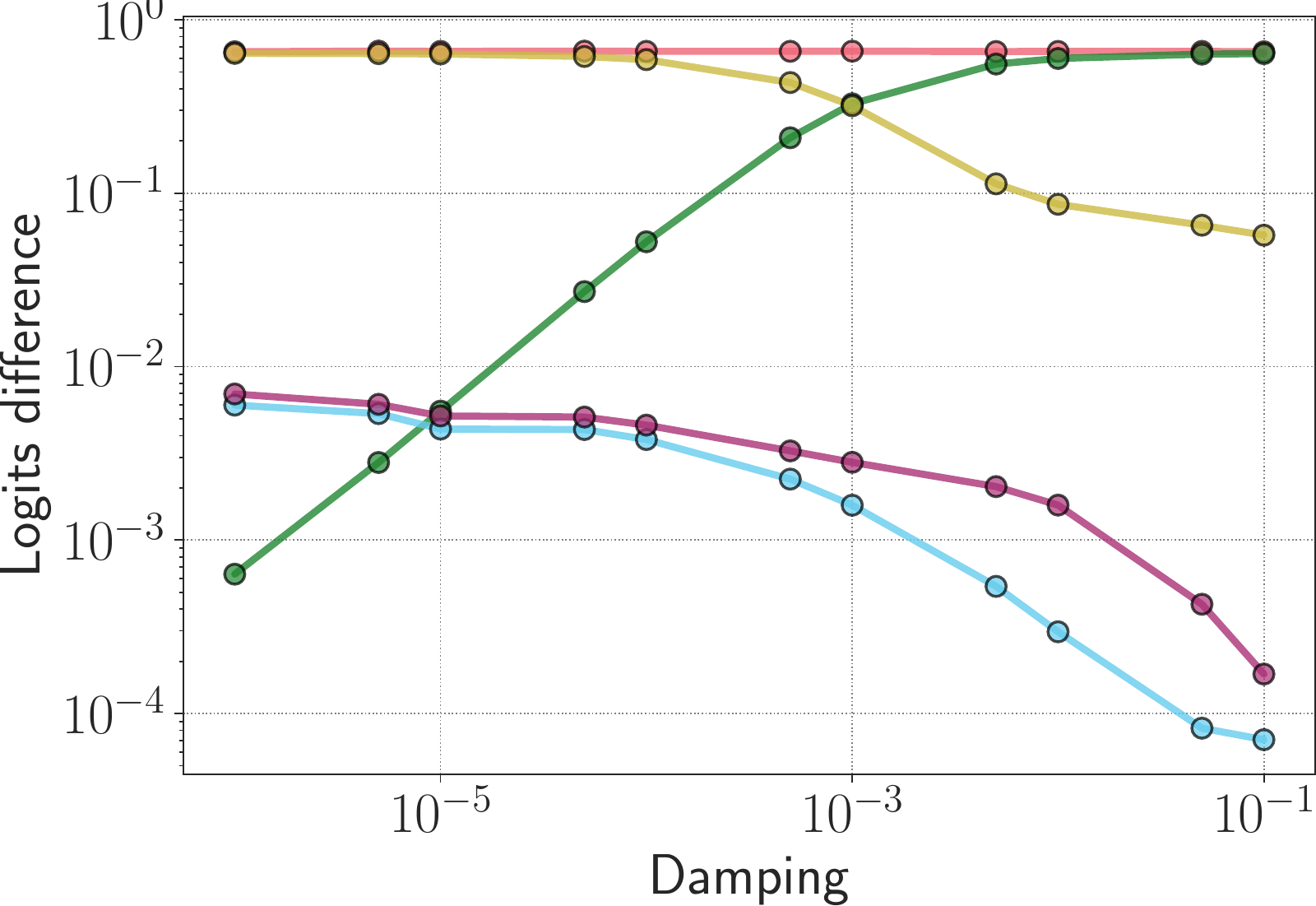}
    \vspace{-0.1cm}
    \caption{{\footnotesize Effect of damping strength}}
    \label{subfig:resnet32-cifar}
\end{subfigure}
\begin{subfigure}[t]{0.32\textwidth}
    \centering
    \includegraphics[height=1.18in]{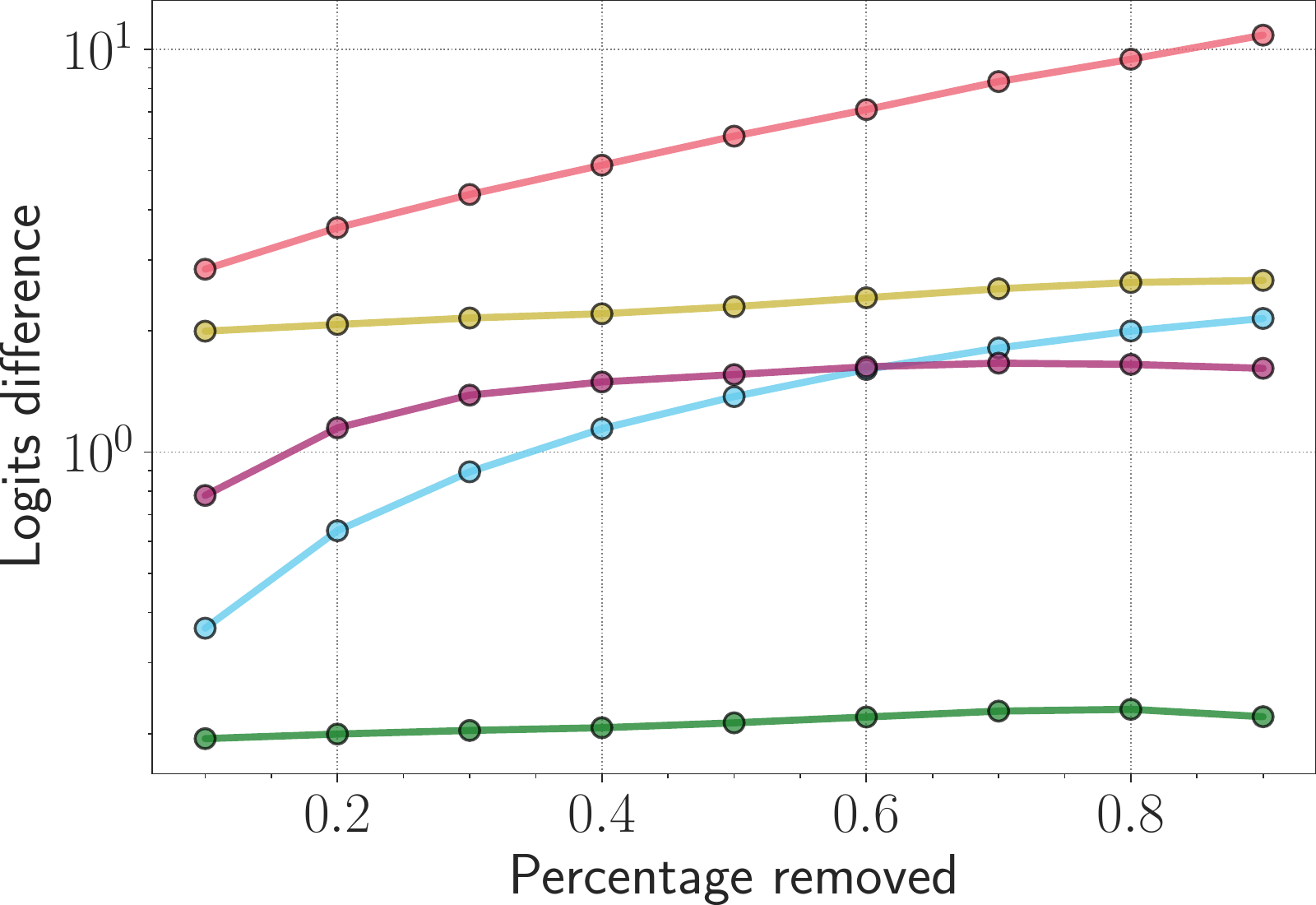}
    \vspace{-0.1cm}
    \caption{{\footnotesize Effect of \% data removed}}
    \label{subfig:transformer-lm1b}
\end{subfigure}
    \vspace{-0.1cm}
\caption{Ablations on how various factors affect the contribution of the gaps and errors to the discrepancy between influence approximation and LOO retraining.}
\vspace{-1.5\baselineskip}
\label{fig:inf_factors} 
\end{figure}

We further analyzed how the contribution of each component
changes in response to changes in network width and depth, training time, weight decay, damping, and the percentage of data removed. We used an MLP trained on 10\% of the MNIST dataset and summarized results in Figure~\ref{fig:inf_factors}.

\parhead{Width and Depth.}  %
We find that as we increase the width of the network, we observe a decrease in the linearization error. This is consistent with previous observations that networks behave more linearly as the width is increased~\citep{lee2019wide}. 
In contrast to the findings from~\citet{basu2020influence}, we did not observe a strong relationship between the contribution of the components and the depth of the network.

\parhead{Training Time.} Unsurprisingly, as we increase the number of training epochs, we observe a decrease in the non-convergence gap. %

\parhead{Weight Decay.} The weight decay allows the training objective to be better conditioned. Consequently, as weight decay increases, the training objective may act more as a strictly convex objective, resulting in a decrease in overall discrepancy for all components.
~\citet{basu2020influence} also found that the alignment between influence functions and LOO retraining increases as weight decay increases. 

\parhead{Damping.} A higher damping term makes linear systems better conditioned, allowing solvers to find accurate solutions in fewer iterations~\citep{demmel1997applied}, thereby reducing the solver error. Furthermore, the higher proximity term keeps the parameters close to $\param^s$, reducing the linearization error. However, increasing the effective proximity penalty directly increases the proximity gap.

\parhead{Percentage of Training Examples Removed.} As we remove more training examples from the dataset the PBRF becomes more non-linear and we observe a sharp increase in the linearization error. The cost landscape is also more likely to change as we remove more training examples, and we observe a corresponding increase in the warm-start gap.

\section{Conclusion}
In this paper, we investigate the sources of the discrepancy between influence functions and LOO retraining in neural networks.
We decompose this difference into five distinct components: the warm-start gap, proximity gap, non-convergence gap, linearization error, and solver error.
We empirically evaluate the contributions of each of these components on a wide variety of architectures and datasets and investigate how they change with factors such as network size and regularization. Our results show that 
the first three components are most responsible for the discrepancy between influence functions and LOO retraining.
We further introduce the proximal Bregman response function (PBRF) to better capture the behavior of influence functions in neural networks.
Compared to LOO retraining, the PBRF is more easily calculated and correlates better with influence functions, meaning it is an attractive alternative gold standard for evaluating influence functions. Although the PBRF may not necessarily align with LOO retraining, it can still be applied in many of the motivating use cases for influence functions. We conclude that influence functions in neural networks are not necessarily ``fragile'', but instead are giving accurate answers to a different question than is normally assumed.
\section*{Acknowledgements}
We would like to thank Pang Wei Koh for helpful discussions. Resources used in this research were provided, in part, by the Province of Ontario, the Government of Canada through CIFAR, and companies sponsoring the Vector Institute (\url{www.vectorinstitute.ai/partners}).

\bibliography{main}
\bibliographystyle{icml2022}

\newpage
\appendix
\section*{Appendix}

This appendix is structured as follows:
\begin{itemize}
    \item In Section~\ref{app:notation}, we provide an overview of the notation we use throughout the paper.
    \item In Section~\ref{app:inf-derivation}, we provide derivations for influence functions, proximal response function, proximal Bregman response function, and linearized proximal response function. 
    \item In Section~\ref{app:decomp}, we present the numerical results shown in Figure~\ref{fig:error_decompse}.
    \item In Section~\ref{app:efficient_computation}, we provide an overview of \texttt{CG} and \texttt{LiSSA} algorithms.
    \item In Section~\ref{app:experiment_details}, we provide experimental details.
    \item In Section~\ref{app:additional_results}, we provide additional experiment results.
\end{itemize}

\clearpage

\section{Table of Notation}
\label{app:notation}

Table~\ref{tab:notation} summarizes notations used in this paper.

\begin{table}[h!]
    \centering
    {\renewcommand{\arraystretch}{1.3}
    \resizebox{\textwidth}{!}{
    \begin{tabular}{cl}
        \toprule
        \textbf{Notation}         & \textbf{Description} \\
        \midrule
        $\mathcal{D}_\text{train}$ & Finite training dataset\\
        $N$ & Number of training examples, $N = |\mathcal{D}_{\text{train}}|$\\  
        $\inp$ & Input of a data point\\
        $\target$ & Target a data point\\
        $\point = (\inp, \target)$ & Data point composed of an input $\inp$ and a target $\target$\\
        $\out = f(\param, \inp)$ & Prediction of the network $f$ parameterized by the parameters $\param$\\
        $\param$ & Model parameters \\
        $\oparam$ & Optimal model parameters on the full training dataset $\mathcal{D}_{\text{train}}$\\
        $\oparam_{-\point}$ & Optimal model parameters on the dataset with a data point $\point$ removed $\mathcal{D}_{\text{train}} - \{\point\}$\\
        $\loss(\out, \target)$ & Loss function (e.g.,~squared error or cross-entropy)\\
        $\cost(\param)$ & Cost function on the full dataset $\mathcal{D}_{\text{train}}$ \\  
        $d$ & Number of model parameters\\  
        $\epsilon$ & Downweighting factor\\  
        $\ucost_{-\point}(\param, \epsilon)$ & Downweighted cost defined as $\ucost_{-\point}(\param, \epsilon) = \cost(\param) - \loss(f(\param, \inp), \target) \epsilon$\\
        $\oparam_{-\point, \epsilon}$ & Optimal model parameters of the downweighted cost $\ucost_{-\point}(\param, \epsilon)$, where $\epsilon$ is fixed\\
        $r^{\star}_{-\point} (\epsilon)$ & Response function\\
        $\mathbf{J}_{\out\oparam}$ & Parameter-output Jacobian at $\oparam$\\
        $\mathbf{H}_{\mathbf{y}}$ & Hessian of the cost with respect to the network outputs\\
        $\lambda$ & Damping strength\\
        $r_{-\point, \text{damp}} ^{\star}(\epsilon)$ & Proximal response function\\
        $\param^s$ & (Possibly non-converged) learned parameters \\
        $\mathcal{D}_{\mathcal{L}^{(i)}} (\mathbf{y}, \mathbf{y}^s)$ & The Bregman divergence $\loss(\mathbf{y}, \mathbf{t}^{(i)}) - \loss(\mathbf{y}^s, \mathbf{t}^{(i)}) -  \nabla_{\mathbf{y}} \loss(\mathbf{y}^s, \mathbf{t}^{(i)})^\top (\mathbf{y} - \mathbf{y}^s )$ \\
        $r^b_{-\point, \text{damp}} (\epsilon)$ & Proximal Bregman response function\\
        $r^b_{-\point, \text{damp}, \text{lin}} (\epsilon)$ & Linearized proximal Bregman response function\\
        $f_{\text{lin}} (\param, \inp)$ & Linearized network outputs with respect to the parameters \\
        $\out^s = f(\param^s, \inp)$ & Prediction of the network $f$ parameterized by the parameters $\param^s$\\
        $\loss_{\text{quad}} (\mathbf{y}, \mathbf{t})$ & Second-order Taylor expansion of the loss around $\mathbf{y}^s$ \\
        \bottomrule
    \end{tabular}
    }}
    \caption{A summary of the notation used in this paper.}
    \label{tab:notation}
\end{table}
\newpage
\section{Derivations}
\label{app:inf-derivation}

\subsection{Influence Function Derivation}
\label{app:der_inf_jacobian}

We provide a derivation of influence functions using the response function. We refer readers to~\citet{van2000asymptotic} and~\citet{koh2017understanding} for a more general derivation of influence functions.

Let $\point = (\inp, \target) \in \trainset$ be a training example we are interested in downweighting. Recall that the downweighted objective is defined as:
\begin{align}
    \ucost_{-\point}(\param, \epsilon) =  \cost(\param) - \loss(f(\param, \inp), \target) \epsilon.
    \label{appeq:dwobj}
\end{align}
We further let $\epsilon_0 = 0$ and $\oparam$ be the optimal parameters such that $\nabla_{\param} \ucost_{-\point} (\oparam, \epsilon_0) = \mathbf{0}$. Here, the optimal parameters $\oparam$ is the solution that minimizes the cost function $\cost(\cdot)$. We assume that the downweighted objective is twice continuously differentiable and strongly convex in the parameters $\param$ at $\epsilon_0$. Note that if we assume the strong convexity of the loss function, the downweighted objective is only guaranteed to be strongly convex when $\epsilon \leq \sfrac{1}{N}$.

By the Implicit Function Theorem, these exists a unique continuously differentiable response function $r^\star_{-\point}\colon \mathcal{U}_0 \to \mathbb{R}^d$ defined on a neighborhood $\mathcal{U}_0$ of $\epsilon_0$ such that $r^{\star}_{-\point} (\epsilon_0) = \oparam$ and:
\begin{align}
    \nabla_{\param} \ucost_{-\point} (r^{\star}_{-\point} (\epsilon), \epsilon) = \mathbf{0}
\end{align}
for all $\epsilon \in \mathcal{U}_0$. By taking the derivative with respect to the downweighting factor $\epsilon$, we get:
\begin{align}
    \mathbf{0} = \frac{\mathrm{d} }{ \mathrm{d} \epsilon} \left( \nabla_{\param} \ucost_{-\point} (r_{-\point}^{\star} (\epsilon), \epsilon) \right)
    =  \nabla_{\param}^2 \ucost_{-\point} (r_{-\point}^{\star} (\epsilon), \epsilon) \frac{\mathrm{d} r^{\star}_{-\point}}{ \mathrm{d} \epsilon} (\epsilon) 
    + \nabla^2_{\param, \epsilon}\ucost_{-\point}(r_{-\point}^{\star} (\epsilon), \epsilon)
    \label{appeq:inf_jacobian}
\end{align}
for all $\epsilon \in \mathcal{U}_0$. The Jacobian of the response function at $\epsilon_0$ can further be expressed as:
\begin{align}
    \mathbf{0} = \left(\nabla^2_{\param} \cost(\oparam) \right) \left( \frac{\mathrm{d} r^{\star}_{-\point}}{ \mathrm{d} \epsilon} \biggr\rvert_{\epsilon = \epsilon_0} \right)
    - \nabla_{\param} \loss(f(\oparam, \inp), \target),
    \label{appeq:inf_jacobian}
\end{align}
where we used these two equalities: 
\begin{align}
    \nabla_{\param}^2 \ucost_{-\point} (r_{-\point}^{\star} (\epsilon_0), \epsilon_0)\ &= \nabla^2_{\param} \cost(\oparam)  \\
    \nabla^2_{\param, \epsilon} \ucost_{-\point} (r_{-\point}^{\star} (\epsilon_0), \epsilon_0) &= -\nabla_{\param} \loss(f(\oparam, \inp), \target)
\end{align}
Rearranging~Eqn.~\ref{appeq:inf_jacobian}, we have:
\begin{align}
    \label{appeq:response_jacobian_full} 
    \frac{\mathrm{d} r^{\star}_{-\point}}{ \mathrm{d} \epsilon} \biggr\rvert_{\epsilon = \epsilon_0} = \left(\nabla^2_{\param} \cost(\oparam) \right)^{-1} \nabla_{\param} \loss(f(\oparam, \inp), \target),
\end{align}
where the Hessian $\nabla^2_{\param} \cost(\oparam)$ is invertible by strong convexity of our downweighted objective at $\epsilon_0$. 
Influence functions approximate the response function with a first-order Taylor expansion at $\epsilon_0$:
\begin{align}
    r_{-\point, \text{lin}}^{\star} (\epsilon) &= r^{\star}_{-\point} (\epsilon_0) +  \frac{\mathrm{d} r^{\star}_{-\point}}{\mathrm{d} \epsilon} \biggr\rvert_{\epsilon = \epsilon_0} (\epsilon - \epsilon_0) = \oparam + (\nabla^2_{\param} \cost(\oparam))^{-1} \nabla_{\param} \loss(f(\oparam, \inp), \target) \epsilon.
\end{align}
To approximate the optimal parameters trained without a data point $\point$, we can substitute $\epsilon = \sfrac{1}{N}$ as follows:
\begin{align}
    r_{-\point, \text{lin}}^{\star} (\sfrac{1}{N}) = \oparam + \frac{1}{N}(\nabla^2_{\param} \cost(\oparam))^{-1} \nabla_{\param} \loss(f(\oparam, \inp), \target).
\end{align}
Influence functions can further approximate the loss at a particular test point $\point_{\text{test}} = (\inp_{\text{test}}, \target_{\text{test}})$ (or test loss) when a training example $\point$ is eliminated from the training set using the chain rule~\citep{koh2017understanding}:
\begin{align}
    \begin{split}
    &\loss(f(r^{\star}_{-\point, \text{lin}} \left( \sfrac{1}{N} \right), \inp_{\text{test}}), \target_{\text{test}}) \\
    &\quad \approx \loss(f(\oparam, \inp_{\text{test}}), \target_{\text{test}}) + \frac{1}{N} \nabla_{\param} \loss(f(\oparam, \inp_{\text{test}}), \target_{\text{test}})^{\top}  \frac{\mathrm{d} r^{\star}_{-\point}}{\mathrm{d} \epsilon} \biggr\rvert_{\epsilon = \epsilon_0}  \\
    &\quad \approx \loss(f(\oparam, \inp_{\text{test}}), \target_{\text{test}}) + \frac{1}{N} \nabla_{\param} \loss(f(\oparam, \inp_{\text{test}}), \target_{\text{test}})^{\top} (\nabla^2_{\param} \cost(\oparam))^{-1} \nabla_{\param} \loss(f(\oparam, \inp), \target).
    \end{split}
    \label{eq:influence_test_loss}
\end{align}

\subsection{Proximal Response Function Derivation}
\label{app:der_damped_response}

Let $\point = (\inp, \target) \in \trainset$ be a training example we are interested in downweighting. Recall that the proximal response function (in Eqn.~\ref{eq:damped_warm_optimum}) is defined as:
\begin{align}
    r_{-\point, \text{damp}} ^{\star}(\epsilon) = \argmin_{\param \in \mathbb{R}^d} \ucost_{-\point} (\param, \epsilon) + \frac{\lambdamp}{2}\|\param - \oparam\|^2,
\end{align}
where $\oparam$ is the optimal parameters that minimize the cost function and $\lambdamp > 0$ is a damping term. Here, we show that influence estimations with a damping term correspond to a first-order Taylor approximation of the proximal response function. Let $\lambdamp > 0$ be some damping term and $\oparam$ be the solution that minimizes the cost function. Let $\epsilon_0 = 0$ and assume that the downweighted objective is convex in the parameters $\param$ at $\epsilon_0$. By the Implicit Function Theorem, we can guarantee the existence of the proximal response function $r_{-\point, \text{damp}} ^{\star}\colon \mathcal{U}_0 \to \mathbb{R}^d$ defined on a neighborhood $\mathcal{U}_0$ of $\epsilon_0$ which satisfies $r^{\star}_{-\point, \text{damp}} (\epsilon_0) = \oparam$ and:
\begin{align}\label{eq:misc21}
    \nabla_{\param} \ucost_{-\point} (r_{-\point, \text{damp}}^{\star}(\epsilon), \epsilon) + \lambdamp \left(r_{-\point, \text{damp}}(\epsilon)^{\star} - \oparam \right) = \mathbf{0}
\end{align}
for all $\epsilon$ in some neighborhood $\mathcal{U}_0$ of $\epsilon_0$. Then, differentiating Eqn.~\ref{eq:misc21} with respect to the downweighting factor $\epsilon$ equates:
\begin{align}
    \mathbf{0} &= \frac{\mathrm{d} }{ \mathrm{d} \epsilon} \left( \nabla_{\param} \ucost_{-\point} (r_{-\point, \text{damp}} ^{\star} (\epsilon), \epsilon) + \lambdamp \left(r_{-\point, \text{damp}}^{\star}(\epsilon) - \oparam \right)\right)\\
    &= \nabla_{\param}^2 \ucost_{-\point} (r_{-\point, \text{damp}} ^{\star} (\epsilon), \epsilon) \frac{\mathrm{d} r_{-\point, \text{damp}} ^{\star}}{ \mathrm{d} \epsilon} (\epsilon) 
    + \nabla^2_{\param, \epsilon}\ucost_{-\point}(r_{-\point, \text{damp}} ^{\star} (\epsilon), \epsilon) + \lambda \frac{\mathrm{d} r_{-\point, \text{damp}} ^{\star}}{\mathrm{d} \epsilon} (\epsilon)
    \label{appeq:inf_jacobian_damp}
\end{align}
for all $\epsilon \in \mathcal{U}_0$.
Evaluating the response Jacobian at $\epsilon_0$ and rearranging the terms in Eqn.~\ref{appeq:inf_jacobian_damp}:
\begin{align}
    \frac{\mathrm{d}r_{-\point, \text{damp}} ^{\star}}{ \mathrm{d} \epsilon} \biggr\rvert_{\epsilon = \epsilon_0} = \left(\nabla^2_{\param} \cost(\oparam) + \lambdamp \mathbf{I}\right)^{-1} \nabla_{\param} \loss(f(\oparam, \inp), \target).
\end{align}
Hence, a first-order Taylor approximation of the proximal response function is equivalent to influence functions with a damping term $\lambda \mathbf{I}$ added:
\begin{align}
    r_{-\point, \text{damp}, \text{lin}}^{\star} (\epsilon) &= r_{-\point, \text{damp}} ^{\star} (\epsilon_0) +  \frac{\mathrm{d} r_{-\point, \text{damp}} ^{\star}}{\mathrm{d} \epsilon} \biggr\rvert_{\epsilon = \epsilon_0} (\epsilon - \epsilon_0) \\
    &= \oparam + (\nabla^2_{\param} \cost(\oparam) + \lambdamp \mathbf{I})^{-1} \nabla_{\param} \loss(f(\oparam, \inp), \target) \epsilon.
\end{align}
When a damping term is used in influence functions, they approximate LOO retraining scheme with the proximity term added to the downweighted objective. 

\subsection{Proximal Bregman Response Function Derivation}
\label{app:suboptimal_response_function}

As opposed to the derivation from Appendix~\ref{app:der_inf_jacobian}, we consider computing influence functions on parameters $\param^s$ that have not necessarily converged. When the parameters have not fully converged, a warm-start retraining with the downweighted objective defined in Eqn.~\ref{appeq:dwobj} would simply minimize the cost function in the first term and reflect the effect of training longer, rather than the effect of removing a training example. 

Let $\point = (\inp, \target) \in \trainset$ be a training example we are interested in downweighting. We assume that the loss function is convex as a function of the network outputs which hold for commonly used loss functions. We replace the cost function in the downweighted objective with a term that penalizes mismatch to the predictions made by the current parameters $\param^s$ and define the proximal Bregman downweighted objective as:
\begin{align}
    \ucost_{-\point}^b (\param, \epsilon) = \frac{1}{N} \sum_{i=1}^N D_{\loss^{(i)}} (f(\param, \inp^{(i)}), f(\param^s, \inp^{(i)})) - \loss(f(\param, \inp), \target) \epsilon + \frac{\lambdamp}{2} \|\param - \param^s\|,
    \label{eq:pbrf_obj}
\end{align}
where $D_{\loss^{(i)}} (\cdot, \cdot)$ is the Bregman divergence defined as:
\begin{align}
    D_{\loss^{(i)}} (\mathbf{y}, \mathbf{y}^s) = \loss(\mathbf{y}, \mathbf{t}^{(i)}) - \loss(\mathbf{y}^s, \mathbf{t}^{(i)}) -  \nabla_{\mathbf{y}} \loss(\mathbf{y}^s, \mathbf{t}^{(i)})^\top (\mathbf{y} - \mathbf{y}^s ).
    \label{appeq:breg}
\end{align}
Because of our convexity assumption, the Bregman divergence term in Eqn.~\ref{eq:pbrf_obj} is non-negative and is $0$ when $\param = \param^s$. Hence, the parameters $\param^s$ is optimal at $\epsilon = 0$ on the proximal Bregman downweighted objective although it is not optimal on the cost function. Accordingly, we define the proximal Bregman response function (PBRF) as follows:
\begin{align}
    r^b_{-\point, \text{damp}} (\epsilon) = \argmin_{\param \in \mathbb{R}^d} \frac{1}{N} \sum_{i=1}^N D_{\loss^{(i)}} (f(\param, \inp^{(i)}), f(\param^s, \inp^{(i)})) - \loss(f(\param, \inp), \target) \epsilon + \frac{\lambdamp}{2} \|\param - \param^s\|,
\end{align}
where we assume that the proximal Bregman downweighted objective is strongly convex at $\epsilon_0$ and the solution to the Bregman downweighted objective is unique.
Letting $\epsilon_0 = 0$, the Bregman response function satisfies $r^b_{-\point, \text{damp}} (\epsilon_0) = \param^s$ and:
\begin{align}
    \nabla_{\param} \ucost_{-\point}^b (r^b_{-\point, \text{damp}} (\epsilon), \epsilon) = \mathbf{0}.
\end{align}
for all downweighting factor $\epsilon$ in some neighborhood of $\epsilon_0$.
By taking the derivative with respect to the downweighting factor $\epsilon$, we get:
\begin{align}
    \mathbf{0} &= \frac{\mathrm{d} }{ \mathrm{d} \epsilon} \left( \nabla_{\param} \ucost_{-\point}^b (r^b_{-\point, \text{damp}} (\epsilon), \epsilon) \right)\\
    &=  \nabla_{\param}^2 \ucost_{-\point}^b (r^b_{-\point, \text{damp}} (\epsilon), \epsilon) \frac{\mathrm{d} r^b_{-\point, \text{damp}}}{ \mathrm{d} \epsilon} (\epsilon) 
    + \nabla^2_{\param, \epsilon}\ucost_{-\point}^b (r^b_{-\point, \text{damp}} (\epsilon), \epsilon).
\end{align}
For linear models, where the parameter-output Jacobian is constant, the Jacobian of the response function at $\epsilon_0$ can further be expressed as:
\begin{align}
    \mathbf{0} = \left(\nabla^2_{\param} \cost(\param^s) + \lambdamp \mathbf{I} \right) \left( \frac{\mathrm{d}r^b_{-\point, \text{damp}}}{ \mathrm{d} \epsilon} \biggr\rvert_{\epsilon = \epsilon_0} \right)
    - \nabla_{\param} \loss(f(\param^s, \inp), \target),
    \label{appeq:inf_jacobian_pbr}
\end{align}
where we used these two equalities: 
\begin{align}
    \nabla_{\param}^2 \ucost_{-\point}^b (r^b_{-\point, \text{damp}} (\epsilon_0), \epsilon_0)\ &= \nabla^2_{\param} \cost(\param^s) + \lambdamp \mathbf{I}  \\
    \nabla^2_{\param, \epsilon} \ucost_{-\point}^b (r^b_{-\point, \text{damp}} (\epsilon_0), \epsilon_0) &= -\nabla_{\param} \loss(f(\param^s, \inp), \target)
\end{align}
Rearranging the terms in Eqn.~\ref{appeq:inf_jacobian_pbr}, the Jacobian of the PBRF at $\epsilon_0$ can be expressed as:
\begin{align}
    \frac{\mathrm{d}r_{-\point} ^{b}}{ \mathrm{d} \epsilon} \biggr\rvert_{\epsilon = \epsilon_0} = \left(\nabla^2_{\param} \cost(\param^s) + \lambdamp \mathbf{I}\right)^{-1} \nabla_{\param} \loss(f(\param^s, \inp), \target).
\end{align}
Note that both Hessian and gradient are computed on the final parameters $\param^s$ instead of the optimal parameters $\oparam$. Hence, influence functions at the non-converged parameters $\param^s$ (with a damping) can be seen as an approximation to the PBRF rather than an approximation to LOO retraining. 

\subsection{Linearized Proximal Bregman Response Function Derivation}
\label{app:lin_response_function_gnh}

We show that the linearized proximal Bregman response function (PBRF) is equivalent to the influence estimation with the Gauss-Newton Hessian approximation and a damping term $\lambda > 0$. Let $\param^s \in \mathbb{R}^d$ be a possibly non-converged learned parameters, $\point = (\inp, \target) \in \trainset$ be a training example we want to downweight, and $\lambdamp > 0$ be a damping term. Recall that the linearized PBRF is defined as:
\begin{align}
\begin{split}
    r^b_{-\point, \text{damp}, \text{lin}} (\epsilon) = \argmin_{\param \in \mathbb{R}^d} \frac{1}{N} &\sum_{i=1}^N D_{\loss^{(i)}_{\text{quad}}} (f_{\text{lin}}(\param, \inp^{(i)}), f(\param^s, \inp^{(i)})) \\
    &\quad- \nabla_{\param}\loss(f(\param^s, \inp), \target)^{\top} \param \epsilon + \frac{\lambdamp}{2} \| \param - \param^s \|^2,
    \label{eq:lin_pbrf2}
\end{split}
\end{align}
where $\loss_{\text{quad}} (\cdot, \cdot)$ and $f_{\text{lin}}(\param, \inp)$ are defined as:
\begin{align}
    \loss_{\text{quad}} (\mathbf{y}, \mathbf{t})= \loss (\mathbf{y}^s, \mathbf{t}) + \nabla_{\mathbf{y}}& \loss(\mathbf{y}^s, \mathbf{t})^{\top} (\mathbf{y} - \mathbf{y}^s) + (\mathbf{y} - \mathbf{y}^s)^{\top} \nabla_{\mathbf{y}}^2 \loss(\mathbf{y}^s, \mathbf{t})(\mathbf{y} - \mathbf{y}^s).\\
    f_{\text{lin}}(\param, \inp^{(i)}) &= f(\param^s, \inp^{(i)}) + \mathbf{J}_{\mathbf{y}^{(i)} \param^s} (\param - \param^s).
\end{align}
Here, $\mathbf{y}^s$ is the prediction of the network parameterized by $\param^s$ and $\mathbf{J}_{\mathbf{y}^{(i)} \param^s}$ is the parameter-output Jacobian of the $i$-th training example. We further assume that the loss function is convex as a function of the network outputs which hold for commonly used loss functions. The first Bregman divergence term in Eqn.~\ref{eq:lin_pbrf2} can be expressed as:
\begin{align}
    D_{\loss^{(i)}_{\text{quad}}} (f_{\text{lin}}(\param, \inp^{(i)}), \mathbf{y}^s) &= \nabla_{\mathbf{y}} \loss(\mathbf{y}^s, \mathbf{t}^{(i)})^{\top} \mathbf{J}_{\mathbf{y}^{(i)} \param^s} (\param - \param^s) \\
    &\quad + (\param - \param^s)^{\top} \mathbf{J}_{\mathbf{y}^{(i)} \param^s}^{\top} \nabla_{\mathbf{y}}^2 \loss(\mathbf{y}^s, \mathbf{t}^{(i)}) \mathbf{J}_{\mathbf{y}^{(i)} \param^s} (\param - \param^s) \\
    &\quad - \nabla_{\mathbf{y}} \loss(\mathbf{y}^s, \mathbf{t}^{(i)})^{\top} \mathbf{J}_{\mathbf{y}^{(i)} \param^s} (\param - \param^s) \\
    &= (\param - \param^s)^{\top} \mathbf{J}_{\mathbf{y}^{(i)} \param^s}^{\top} \nabla_{\mathbf{y}}^2 \loss(\mathbf{y}^s, \mathbf{t}^{(i)}) \mathbf{J}_{\mathbf{y}^{(i)} \param^s} (\param - \param^s).
\end{align}
Now, taking the gradient of linearized PBRF objective with respect to the parameters $\param$ and setting it equal to $\mathbf{0}$, we get:
\begin{align}
\begin{split}
    \mathbf{0} &= \frac{1}{N} \sum_{i=1}^N \left(  \mathbf{J}_{\mathbf{y}^{(i)} \param^s}^{\top} \nabla_{\mathbf{y}}^2 \loss(\mathbf{y}^s, \mathbf{t}^{(i)}) \mathbf{J}_{\mathbf{y}^{(i)} \param^s} (\param - \param^s) \right) - \nabla_{\param}\loss(f(\param^s, \inp), \target)  \epsilon + \lambdamp(\param - \param^s) \\
    &= \mathbf{J}_{\mathbf{y} \param^s}^{\top} \mathbf{H}_\mathbf{y}^s \mathbf{J}_{\mathbf{y} \param^s} (\param - \param^s ) -\nabla_{\param}\loss(f(\param^s, \inp), \target) \epsilon + \lambdamp(\param - \param^s),
\end{split}
\end{align}
where $\mathbf{H}_{\mathbf{y}}^s$ is the Hessian of the loss with respect to the network outputs evaluated at $\mathbf{y}^s$.
Rearranging the terms, we get:
\begin{align}
    \oparam_{\text{lin, PBRF}} = \param^s + \left(\mathbf{J}_{\mathbf{y} \param^s}^{\top} \mathbf{H}_\mathbf{y}^s \mathbf{J}_{\mathbf{y} \param^s} + \lambdamp \mathbf{I} \right)^{-1} \nabla_{\param}\loss(f(\param^s, \inp), \target)\epsilon,
    \label{appeq:lin_pbrf_sol}
\end{align}
where $\oparam_{\text{lin, PBRF}}$ is the optimal solution to the linearized PBRF objective. Note that the Gauss-Newton Hessian (GNH) $\mathbf{G}^s = \mathbf{J}_{\mathbf{y} \param^s}^{\top} \mathbf{H}_\mathbf{y}^s \mathbf{J}_{\mathbf{y} \param^s}$ is positive semidefinite (assuming that the loss function is convex as a function of network outputs) and an addition of damping term guarantees the invertibility. Therefore, Eqn.~\ref{appeq:lin_pbrf_sol} is equivalent to the influence estimation with the GNH approximation and a damping term $\lambdamp$.

\section{Influence Misalignment Decomposition Table}
\label{app:decomp}

In Table~\ref{tab:results2}, we present the numerical results shown in Figure~\ref{fig:error_decompse}.

\section{Efficient \texttt{iHVP} Computation}
\label{app:efficient_computation}

One of the major challenges in applying influence functions to neural networks (Eqn.~\ref{eq:first_order_response}) is that they involve the computation of an inverse-Hessian vector product ($\texttt{iHVP}$). However, for large networks, computing \texttt{iHVP}s exactly via storing and inverting the Hessian is intractable. To circumvent this,~\citet{koh2017understanding} consider alternative methods for approximating \texttt{iHVP}s, namely, the method of conjugate gradients (\texttt{CG})~\citep{martens2010deep} and the Linear time Stochastic Second-Order Algorithm (\texttt{LiSSA})~\citep{agarwal2016second}.

\parhead{Conjugate Gradients.} Given a positive-definite damped Hessian $\nabla_{\param}^2\cost(\param^s) + \lambda\mathbf{I}$ (where $\param^s$ are the potentially-suboptimal parameters at which the Hessian is taken and $\lambda > 0$ is a damping factor) and vector $\mathbf{v} \in \mathbb{R}^d$, \texttt{CG} arrives at the \texttt{iHVP} by solving an equivalent convex quadratic optimization problem:
\begin{equation}
    (\nabla_{\param}^2\cost(\param^s) + \lambda\mathbf{I})^{-1}\mathbf{v}  = \argmin_{\mathbf{t} \in \mathbb{R}^d}\frac{1}{2}\mathbf{t}^\top(\nabla_{\param}^2\cost(\param^s) + \lambda\mathbf{I})\mathbf{t} - \mathbf{v}^\top \mathbf{t}.
\end{equation}
The \texttt{CG} algorithm starts with an initial guess $\mathbf{v}_0 \in \mathbb{R}^d$ and iteratively updates it, with the bottleneck at each step being an $O(Nd)$ Hessian-vector product. Although an exact solution is only guaranteed after $d$ \texttt{CG} iterations, in practice,~\citet{koh2017understanding} use truncated \texttt{CG} with fewer iterations and achieve a sufficiently close approximation.

\parhead{LiSSA.} The \texttt{LiSSA} algorithm approximates the \texttt{iHVP} using a truncated Neumann series. Given a positive-definite damped Hessian $\nabla_{\param}^2\cost(\param^s) + \lambda\mathbf{I}$ and vector $\mathbf{v} \in \mathbb{R}^d$, we have: 
\begin{equation}\label{eq:truncated newman}
    (\nabla_{\param}^2\cost(\param^s) + \lambda\mathbf{I})^{-1}\mathbf{v}
    \approx \sigma^{-1}\sum_{t = 1}^{T}((1 - \sigma^{-1}\lambda)\mathbf{I} - \sigma^{-1}\nabla_{\param}^2\cost(\param^s))^t\mathbf{v},
\end{equation}
which becomes exact as the recursion depth $T$ approaches $\infty$. Here, $\sigma > 0$ is a scaling hyperparameter that is chosen sufficiently large to ensure convergence of the series. Eqn.~\ref{eq:truncated newman} can be recursively computed over $T$ iterations, with each step requiring an $O(Nd)$ Hessian-vector product. In practice, the computation is further optimized by estimating $\nabla_{\param}^2\cost(\param^s)$ using a randomly-sampled batch $\mathcal{B} \subseteq \trainset$ of size $|\mathcal{B}| \ll N$, so that the Hessian-vector product is reduced to $O(d)$ cost. Then, to accommodate for the added stochasticity, the \texttt{iHVP} is estimated by averaging Eqn.~\ref{eq:truncated newman} over $R$ trial repeats. Hence, the \texttt{LiSSA} algorithm estimates:
\begin{equation}
    (\nabla_{\param}^2\cost(\param^s) + \lambda\mathbf{I})^{-1}\mathbf{v}
    \approx \frac{1}{R}\sum_{r = 1}^R\left(\sigma^{-1}\sum_{t = 1}^{T}((1 - \sigma^{-1}\lambda)\mathbf{I} - \sigma^{-1}\nabla_{\param}^2\cost^{(r, t)}(\param^s))^t\mathbf{v}\right),
\end{equation}
where $\cost^{(r, t)}(\param^s)$ is the average loss over the $(r, t)$-th sampled batch of data.

\parhead{The $s_{\text{test}}$ trick.} Finally, we note that another simple trick can be made when using influence functions to predict the change in test loss at a particular test point (Eqn.~\ref{eq:influence_test_loss}). It is often the case that we wish to compute the influence scores for the pairwise interactions of the entire training dataset on the loss at a comparatively smaller number $N_{\text{test}} \ll N$ of test points. Since the Hessian is symmetric, the order of multiplication in the second term of Eqn.~\ref{eq:influence_test_loss} can be permuted as follows:     
\begin{equation}\label{eq:stest trick}
    \frac{1}{N}\mathbf{v}_{\text{test}}^\top \left[ (\nabla^2_{\param} \cost(\param^s) + \lambda \mathbf{I})^{-1}  \mathbf{v}\right]
    =  \frac{1}{N} \mathbf{v}^{\top} \left[
    (\nabla^2_{\param} \cost(\param^s) + \lambda \mathbf{I})^{-1} \mathbf{v}_{\text{test}}\right],
\end{equation}
where $\mathbf{v} = \nabla_{\param} \loss(f(\param^s, \inp), \target)$  and $\mathbf{v}_{\text{test}} = \nabla_{\param} \loss(f(\param^s, \inp_{\text{test}}), \target_{\text{test}})$. 
This means that we can precompute $\mathbf{s}_{\text{test}} = (\nabla^2_{\param} \cost(\param^s) + \lambda \mathbf{I})^{-1} \mathbf{v}_{\text{test}}$ over all $N_{\text{test}}$ test points of interest, and then cheaply compute influence scores over all $N$ training points by simply taking dot products of the form $\mathbf{v}^\top \mathbf{s}_{\text{test}}$. We refer readers to~\citet{koh2017understanding} for details. 
\section{Experimental Details}
\label{app:experiment_details}

\subsection{Computing Environment}
All experiments were implemented using the PyTorch~\citep{NEURIPS2019_9015} and JAX~\citep{jax2018github} frameworks and we ran all experiments on NVIDIA P100 GPUs.

\subsection{Experiment Set-up}

In all experiments, we first trained the base network (initialized with some $\param^0$) with the entire dataset for $K$ epochs to obtain the base parameters $\param^s$. We then select 20 random data points $\point_t \in \mathcal{D}_{\text{train}}$ from the training dataset and computed the additional parameters for each $t$ as follows: 
\begin{enumerate}
    \item \textbf{Cold optimum.} We retrained the network for $K + \sfrac{K}{2}$ epochs with the initialization $\param^0$. To minimize the effect of stochasticity in stochastic gradient-based optimizer, we further used the same batch order used for training the base network for the first $K$ epochs. 
    \item \textbf{Warm optimum.} We retrained the network for $\sfrac{K}{2}$ epochs with the initialization $\param^s$.
    \item \textbf{Proximal warm optimum.} We retrained the network for $\sfrac{K}{2}$ epochs with the initialization $\param^s$ using the objective defined in Eqn.~\ref{eq:damped_warm_optimum}.
    \item \textbf{Proximal Bregman warm optimum.} We retrained the network for $\sfrac{K}{2}$ epochs with the initialization $\param^s$ using the objective defined in Eqn.~\ref{eq:pbrf}.
    \item \textbf{Linearized proximal Bregman warm optimum.} We retrained the network for $\sfrac{K}{2}$ epochs with the initialization $\param^s$ using the objective defined in Eqn.~\ref{eq:lin_pbrf}.
    \item \textbf{Influence estimation.} We used the \texttt{LiSSA} algorithm on the base parameters $\param^s$ with the Gauss-Newton Hessian approximation (Eqn.~\ref{eq:inf_gn_approx}).
\end{enumerate}

The ``warm-start gap'' refers to the discrepancy between cold-start and warm-start optima. The ``proximity gap'' refers to the discrepancy between warm-start and proximal warm-start optima. The ``non-convergence gap'' denotes the discrepancy between proximal warm-start and proximal Bregman warm-start optima. The ``linearization error'' represents the discrepancy between proximal Bregman warm-start and linearized proximal Bregman warm-start optima. Lastly, the ``solver error'' denotes the discrepancy between linearized proximal Bregman warm optimum and influence estimation with the \texttt{LiSSA} algorithm.

We treated the scaling in the \texttt{LiSSA} algorithm as a separate hyperparameter~\citep{koh2017understanding} and tuned the scaling in the range \{10, 25, 50, 100, 150, 200, 250, 300, 400, 500\} so that the algorithm converges.

\subsection{Influence Misalignment Decomposition}

\parhead{Logistic Regression.} We used Cancer and Diabetes classification datasets from the UCI collection~\citep{Dua2019}. In training, we normalized the input features to have zero mean and unit variance. We trained the model using \texttt{L-BFGS} with $L_2$ regularization of 0.01 and damping term of $\lambda = 0.001$. 

\parhead{Multilayer Perceptron.} For regression experiments, we used 2-hidden layer MLP with 128 hidden units. For both Concrete and Energy datasets, we normalized the input features and targets to have a zero mean and unit variance. For classification experiments with 10\% of MNIST~\citep{deng2012mnist} and FashionMNIST~\citep{xiao2017fashion} datasets, we used 2-hidden layer MLP with a hidden unit dimension of 1024. For both regression and classification experiments, we used a batch size of 128 and trained the base network for 1000 epochs using SGD. While we did not use any $L_2$ regularization, we set the damping strength to $\lambda = 0.001$.

We conducted the hyperparameter searches over the learning rates for the base model, making choices based on the final validation loss. We swept over the learning rates \{1.0, 0.3, 0.1, 0.03, 0.01, 0.003, 0.001\}. We used a learning rate decayed by a factor of 10 for computing PBRF and linearized PBRF. We set the recursion depth to 5000 and the number of repeat to 5 for the \texttt{LiSSA} algorithm.

\parhead{Autoencoder.} We used the same experimental set-up from~\citet{martens2015optimizing}. The loss function was the binary cross-entropy and the $L_2$ regularization with strength $5 \cdot 10 ^{-5}$ was added to the cost function. The layer widths for the autoencoder were [784, 1000, 500, 250, 30, 250, 500, 1000, 784] and we used sigmoid activation functions. We trained the network for 1000 epochs on the full MNIST dataset with SGDm (SGD with momentum) and set the batch size to 1024. The damping term was set to $\lambda = 0.001$.

We conducted the hyperparameter searches for the base model making choices based on the final validation loss. We kept the momentum to $0.9$ and swept over the learning rates 1, 0.3, 0.1, 0.03, 0.01, 0.003, 0.001. We used a learning rate decayed by a factor of 10 for computing PBRF and linearized PBRF. We set the recursion depth to 10000 and the number of repeat to 5 for the \texttt{LiSSA} algorithm.

\parhead{Convolution Neural Networks.} We trained LeNet~\citep{lecun1998gradient}, AlexNet~\citep{krizhevsky2012imagenet}, VGG13~\cite{simonyan2014very}, and ResNet-20~\citep{he2015deep} on 10\% of MNIST dataset and the full CIFAR10~\citep{Krizhevsky2009learning} dataset. For the MNIST experiment, we kept the learning rate fixed, while for CIFAR10 experiment, we decayed the initial learning rate by a factor of 5 at epochs 60, 120, 160. We used $L_2$ regularization of $5 \cdot 10^{-4}$ and a damping factor of $\lambda = 0.001$. For both datasets, we trained the base network for 200 epochs with the batch size of 128.

We used 10\% of the MNIST dataset for the test loss correlation experiments in Table~\ref{table:corr_inf} and computed the approximated test loss with Eqn.~\ref{eq:influence_test_loss} on randomly selected test examples. We conducted the hyperparameter searches for the base model making choices based on the final validation accuracy. We fixed the momentum to $0.9$ and swept over the initial learning rates \{1.0, 0.3, 0.1, 0.03, 0.01, 0.003, 0.001\}. We used a learning rate decayed by a factor of 10 for computing PBRF and linearized PBRF. We set the recursion depth to 10000 and the number of repeat to 5 for the \texttt{LiSSA} algorithm.

\parhead{Transformer.} We trained a 2-layer Transformer language model on the Penn Treebank (PTB) dataset~\citep{marcus1993building}. The number of hidden dimensions was set to $256$ and the number of attention heads was set to $2$. We trained the model with Adam for 10 epochs. We set the batch size to $20$ and a damping term of $\lambda = 0.01$. We conducted the hyperparameter searches for the base model making choices based on the final validation perplexity. We swept over the learning rates \{0.03, 0.01, 0.003, 0.001, 0.0003, 0.0001\} for training the base model and set the recursion depth to 5000 and the number of repeat to 5 for the \texttt{LiSSA} algorithm.

\subsection{Factors in Influence Misalignment}

The experiment in Section~\ref{sec:factors_exp} was conducted using 10\% of the MNIST dataset. We trained 2-hidden layer MLP composed of 1024 hidden units for 1000 epochs using SGD. We used the batch size of 128 and conducted the hyperparameter searches for each model making choices based on the final validation loss. We swept over the learning rates \{1.0, 0.3, 0.1, 0.03, 0.01, 0.003, 0.001, 0.0003, 0.0001\}. 

To see how the gaps and errors change as we increase the width of the network, we repeated the experiments with all widths in a set \{16, 32, 64, 128, 256, 512, 1024, 2048, 4096, 8192\} while keeping the depth fixed to 2. To see the effect of increase in the depth of the network, we fixed the width to 1024 and computed gaps and errors with depth \{1, 3, 5, 7, 9, 11, 13, 15\}.

While keeping all configurations the same (2-hidden layer MLP with 1024 hidden units), we computed the decomposition by changing the total number of epochs to train the base model in the range \{100, 500, 1000, 3000, 5000, 7000, 9000\}, changing the strength of the weight decay \{0.5, 0.1, 0.05, 0.01, 0.005, 0.001, 0.0005, 0.0001, 0.00005, 0.00001\}, and changing the strength of the damping term in the range \{0.5, 0.1, 0.05, 0.01, 0.005, 0.001, 0.0005, 0.0001, 0.00005, 0.00001, 0.000005, 0.000001\}. Lastly, we modified the downweighted objective to remove a group of training examples (rather than a single training example) and altered the percentage we remove the training dataset in range \{0.1, 0.2, 0.3, 0.4, 0.5, 0.6, 0.7, 0.8, 0.9\}.

\section{Additional Results}
\label{app:additional_results}

\begin{table}[t]
    \centering
    \resizebox{\columnwidth}{!}{%
    \begin{tabular}{llccccc}
        \toprule
        \textbf{Model} & \textbf{Dataset} & \textbf{Warm-Start} & \textbf{Proximity} & \textbf{Non-Convergence} & \textbf{Linearization} & \textbf{Solver}\\
        \midrule 
        \multirow{4}{*}{MLP} & Concrete & 0.079 $\pm$ 0.008 & 0.002 $\pm$ 0.000 & \textbf{0.082} $\pm$ 0.001 & 0.000 $\pm$ 0.000 & 0.017 $\pm$ 0.026\\
        & Energy & \textbf{0.016} $\pm$ 0.001 & 0.001 $\pm$ 0.000 & \textbf{0.016} $\pm$ 0.001 & 0.000 $\pm$ 0.000 & 0.000 $\pm$ 0.000 \\
        & MNIST & \textbf{0.208} $\pm$ 0.732 & 0.006 $\pm$ 0.024 & 0.211 $\pm$ 0.745 & 0.000 $\pm$ 0.000 & 0.003 $\pm$ 0.006\\
        & FashionMNIST & \textbf{2.968} $\pm$ 0.545 & 0.385 $\pm$ 0.028 & 0.408 $\pm$ 0.033 & 0.000 $\pm$ 0.001 & 0.009 $\pm$ 0.014\\
        \bottomrule
    \end{tabular}
    }
    \caption{Decomposition of the discrepancy between influence functions (without the Gauss-Newton Hessian approximation) and LOO retraining into (1) warm-start gap, (2) proximal gap, (3) non-convergence gap, (4) linearization error, and (5) solver error for each model and dataset. Different from Table~\ref{tab:results2}, we computed influence functions \textbf{without} the Gauss-Newton Hessian approximation. The size of each component is measured by the $L_2$ distance between the networks' outputs on the training dataset.}
    \label{tab:results_hvp}
\end{table}

\begin{table}[t]
    \centering
    \resizebox{\columnwidth}{!}{%
    \begin{tabular}{llccccc}
        \toprule
        \textbf{Model} & \textbf{Dataset} & \textbf{Warm-Start} & \textbf{Proximity} & \textbf{Non-Convergence} & \textbf{Linearization} & \textbf{Solver}\\
        \midrule 
        \multirow{2}{*}{LR} & Cancer & 0.000 $\pm$ 0.000 & 0.000 $\pm$ 0.000 & 0.000 $\pm$ 0.000 & 0.000 $\pm$ 0.000 & 0.000 $\pm$ 0.000\\
        & Diabetes & 0.000 $\pm$ 0.000 & 0.000 $\pm$ 0.000 & 0.000 $\pm$ 0.000 & 0.000 $\pm$ 0.000 & 0.000 $\pm$ 0.000\\
        \midrule
        \multirow{4}{*}{MLP} & Concrete & 0.079 $\pm$ 0.008 & 0.002 $\pm$ 0.000 & \textbf{0.082} $\pm$ 0.001 & 0.000 $\pm$ 0.000 & 0.001 $\pm$ 0.002\\
        & Energy & \textbf{0.016} $\pm$ 0.001 & 0.001 $\pm$ 0.000 & \textbf{0.016} $\pm$ 0.001 & 0.000 $\pm$ 0.000 & 0.000 $\pm$ 0.000 \\
        & MNIST & \textbf{0.208} $\pm$ 0.732 & 0.006 $\pm$ 0.024 & 0.211 $\pm$ 0.745 & 0.000 $\pm$ 0.000 & 0.001 $\pm$ 0.004\\
        & FashionMNIST & \textbf{2.968} $\pm$ 0.545 & 0.385 $\pm$ 0.028 & 0.408 $\pm$ 0.033 & 0.000 $\pm$ 0.001 & 0.007 $\pm$ 0.010\\
        \midrule
        Autoencoder & MNIST & \textbf{20.743} $\pm$ 0.522 & 14.330 $\pm$ 0.184 & 11.409 $\pm$ 0.430 & 0.000 $\pm$ 0.000 & 0.307 $\pm$ 0.088\\
        \midrule
        LeNet & \multirow{4}{*}{MNIST} & \textbf{7.434} $\pm$ 1.068 & 5.393 $\pm$ 0.547 & 3.748 $\pm$ 0.324 & 0.000 $\pm$ 0.001 & 0.001 $\pm$ 0.001 \\
        AlexNet & & \textbf{13.403} $\pm$ 1.289 & 0.001 $\pm$ 0.000 & 0.118 $\pm$ 0.000 & 0.000 $\pm$ 0.000 & 0.013 $\pm$ 0.002  \\
        VGG13 & & 7.389 $\pm$ 0.744 & 4.872 $\pm$ 1.052 & \textbf{8.601} $\pm$ 0.478 & 0.000 $\pm$ 0.000 & 0.001 $\pm$ 0.001  \\
        ResNet-20 & & 4.433 $\pm$ 0.059 & 4.061 $\pm$ 0.157 & \textbf{4.940} $\pm$ 0.155 & 0.001 $\pm$ 0.000 & 0.002 $\pm$ 0.001 \\
        \midrule
        LeNet & \multirow{4}{*}{CIFAR10} & \textbf{10.668} $\pm$ 0.162 & 6.520 $\pm$ 0.442 & 5.032 $\pm$ 0.799 & 0.001 $\pm$ 0.000 & 0.000 $\pm$ 0.000 \\
        AlexNet & & \textbf{7.530} $\pm$ 0.233 & 5.956 $\pm$ 0.102 & 2.864 $\pm$ 0.367 & 0.000 $\pm$ 0.000 & 0.000 $\pm$ 0.000  \\
        VGG13 & & \textbf{8.410} $\pm$ 1.926 & 6.279 $\pm$ 0.708 & 6.031 $\pm$ 0.660 & 0.000 $\pm$ 0.000 & 0.073 $\pm$ 0.161 \\
        ResNet-20 & & \textbf{5.827} $\pm$ 0.152 & 4.435 $\pm$ 0.669 & 3.280 $\pm$ 0.429 & 0.000 $\pm$ 0.000 & 0.003 $\pm$ 0.001 \\
        \midrule
        Transformer & PTB & 57.926 $\pm$ 5.055 & 30.756 $\pm$ 0.673 & \textbf{61.675} $\pm$ 1.042 & 5.002 $\pm$ 1.556 & 3.316 $\pm$ 2.630\\
        \bottomrule
    \end{tabular}
    }
    \caption{Decomposition of the discrepancy between influence functions and LOO retraining into (1) warm-start gap, (2) proximal gap, (3) non-convergence gap, (4) linearization error, and (5) solver error for each model and dataset. The size of each component is measured by the $L_2$ distance between the networks' outputs on the training dataset.}
    \label{tab:results2}
\end{table}

\begin{table}[t]
    \centering
    \footnotesize
    \begin{tabular}{@{}ccccccc@{}}
    \toprule
    \textbf{Dataset} & \multicolumn{2}{c}{\textbf{Cold-Start}} & \multicolumn{2}{c}{\textbf{Warm-Start}} & \multicolumn{2}{c}{\textbf{PBRF}} \\ \cmidrule(l){2-7} 
     & P & S & P & S & P & S \\ \midrule
    Concrete & -0.11 & 0.12 & 0.09 & 0.11 & \textbf{0.93}  &  \textbf{0.94}\\ 
    Energy & 0.03 & 0.04 & 0.09 & 0.13 & \textbf{0.97}  &  \textbf{0.91}\\ 
    MNIST & -0.10 & 0.01 & 0.22 & 0.35 & \textbf{0.98} & \textbf{0.91} \\
    FashionMNIST & 0.16 & 0.00 & 0.06 & 0.07 & \textbf{0.90} & \textbf{0.92} \\ \bottomrule
    \end{tabular}
    \vspace{-0.2cm}
    \caption{Comparison of test loss differences computed by influence function (\textbf{with} the Gauss-Newton Hessian approximation), cold-start retraining, warm-start retraining, and PBRF. We show Pearson (P) and Spearman rank-order (S) correlation when compared to influence estimates.
    }
    \label{tab:results_hvp2}
\end{table}

\begin{table}[t]
    \centering
    \footnotesize
    \begin{tabular}{@{}ccccccc@{}}
    \toprule
    \textbf{Dataset} & \multicolumn{2}{c}{\textbf{Cold-Start}} & \multicolumn{2}{c}{\textbf{Warm-Start}} & \multicolumn{2}{c}{\textbf{PBRF}} \\ \cmidrule(l){2-7} 
     & P & S & P & S & P & S \\ \midrule
    Concrete & -0.14 & 0.32 & 0.24 & 0.37 & \textbf{0.85}  &  \textbf{0.71}\\ 
    Energy & -0.01 & -0.24 & 0.44 & 0.57 & \textbf{0.95}  &  \textbf{0.83}\\ 
    MNIST & -0.16 & -0.12 & 0.21 & -0.14 & \textbf{0.98} & \textbf{0.81} \\
    FashionMNIST & -0.19 & 0.01 & 0.05 & 0.24 & \textbf{0.91} & \textbf{0.88} \\ \bottomrule
    \end{tabular}
    \vspace{-0.2cm}
    \caption{Comparison of test loss differences computed by influence function (\textbf{without} the Gauss-Newton Hessian approximation), cold-start retraining, warm-start retraining, and PBRF. We show Pearson (P) and Spearman rank-order (S) correlation when compared to influence estimates.
    }
    \label{tab:results_hvp3}
\end{table}

\subsection{Influence Functions without the Gauss-Newton Hessian Approximation}
\label{sec:hessian_approx}

In all experiments, we computed influence functions using the Gauss-Newton Hessian (GNH) approximation. As the previous error analysis was conducted without the GNH approximation~\citep{basu2020influence}, we repeated a subset of our experiments without the GNH approximation (using the Hessian matrix). The results are summarized in Table~\ref{tab:results_hvp}. 

As the warm-start gap, proximity gap, non-convergence gap, and linearization error do not depend on the way influence estimates are computed, these numerical values are identical to the results in Table~\ref{tab:results2}. However, as the linearized PBRF optimum is equivalent to influence estimations with the GNH approximation, the solver error slightly increased when we computed influence functions with the Hessian for all datasets. Nevertheless, the solver error is still significantly lower than other decomposition terms.

Furthermore, we investigated how the test loss difference on randomly selected test examples approximated by influence functions correlates with the actual value computed using cold-start retraining, warm-start retraining, and the PBRF. Table~\ref{tab:results_hvp2} shows the correlations with influence approximations using the GNH approximation. While influence estimates do not accurately predict the effect of retraining the model, they closely align the values obtained by the PBRF. Moreover, we conducted the same experiment with influence approximations without the GNH approximation (using the Hessian matrix) and show the results in Table~\ref{tab:results_hvp3}. Similar to the previous results, the test loss differences predicted by influence functions align with the value obtained by the PBRF while failing to capture the effect of retraining the model for both regression and classification datasets. Hence, in both settings, the PBRF better captures the behavior of influence functions than LOO retraining.

\subsection{Two-Stage LOO Retraining: An Alternative Method for PBRF computation}
\label{app:altern}

As discussed in Section~\ref{sec:non_converge}, when the network has not fully converged, the LOO retraining simply reflects the effect of training the network for a longer period of time, which does not correctly reflect the effect of removing a data point from the training set (a question that influence functions aim to answer). This difference yields the discrepancy between LOO retraining and influence estimates. We further introduced the PBRF objective which penalizes the LOO retraining with both function- and weight-space discrepancy terms and showed that the PBRF better reflects the question influence functions try to answer. 

Here, we discuss an alternative method to capture the behaviour of influence functions by performing two separate LOO retrainings, one with the full dataset and the other one with the removed data point. The first LOO retraining reflects the effect of training longer with the full dataset and the second LOO retraining reflects the effect of training longer with a removed data point. The difference between LOO retrainings in parameter space can approximate the effect of removing a data point by neglecting the effect of longer training. Such difference can be added to the current parameters $\param^s$. This two-stage LOO retraining method can be interpreted as removing $\mathcal{O}(1)$ terms when the network has not fully converged. 

\begin{table}[t]
    \centering
    \footnotesize
    \begin{tabular}{@{}ccccccc@{}}
    \toprule
    \textbf{Dataset} & \multicolumn{2}{c}{\textbf{Two-Stage LOO}} & \multicolumn{2}{c}{\textbf{Warm-Start}} & \multicolumn{2}{c}{\textbf{PBRF}} \\ \cmidrule(l){2-7} 
     & P & S & P & S & P & S \\ \midrule
    Concrete & 0.52 & 0.61 & 0.24 & 0.37 & \textbf{0.85}  &  \textbf{0.71}\\ 
    Energy & 0.92 & 0.75 & 0.44 & 0.57 & \textbf{0.95}  &  \textbf{0.83}\\ 
    MNIST & 0.73 & 0.53 & 0.21 & -0.14 & \textbf{0.98} & \textbf{0.81} \\
    FashionMNIST & 0.69 & 0.52 & 0.05 & 0.24 & \textbf{0.91} & \textbf{0.88} \\ \bottomrule
    \end{tabular}
    \caption{Comparison of test loss differences computed by influence function, warm-start retraining, PBRF, and two-stage LOO retraining on MNIST dataset. We show Pearson (P) and Spearman rank-order (S) correlation when compared to influence estimates.
    }
    \label{tab:results_corr_with_two_stage}
\end{table}

We repeat the experiment from Appendix~\ref{sec:hessian_approx} with the proposed two-stage LOO retraining. The results are shown in Table~\ref{tab:results_corr_with_two_stage}. While the two-stage LOO retraining shows worse correlation with influence estimates compared to the PBRF objective, it has significantly higher correlation with influence estimates compared to the warm-start retraining. Although the two-stage LOO retraining requires retraining the network twice, it can be seen as a better reflection to what influence functions compute in neural networks.

\subsection{Mislabelled Examples Detection}
\label{app:mislabelled}

\begin{figure}[h]
    \centering
    \includegraphics[width=0.45\textwidth]{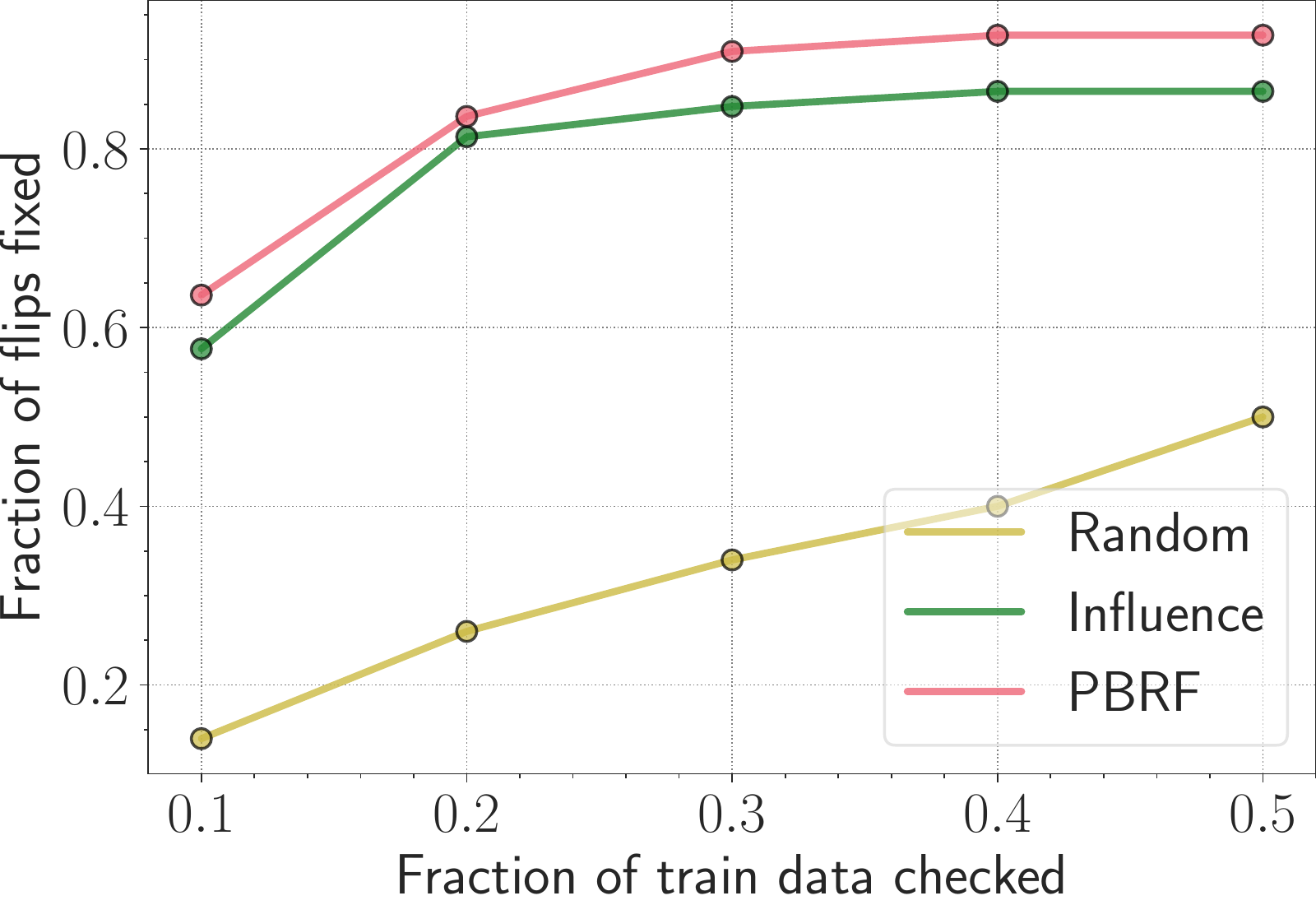}
    \small
    \caption{Effectiveness of PBRF and influence functions on fixing mislabeled training examples on corrupted MNIST dataset. We examine the fraction of the training data to fix the mislabelling while prioritizing the data examples with higher influential scores produced by PBRF and influence functions. The PBRF and influence functions help detect mislabelled examples.}
    \label{fig:fix}
\end{figure}

While the PBRF may not necessarily align with LOO retraining because of warm-start, proximity, and non-convergence gaps, the motivating use cases for influence functions typically do not rely on exact LOO retraining. Hence, the PBRF can be used instead of LOO retraining for many tasks, such as identifying influential or mislabelled examples. We conducted an additional experiment to verify that the PBRF (and influence function) is still helpful in detecting mislabelled training data points.

We used 10\% of the MNIST dataset and randomly corrupted 10\% of the training examples by assigning random labels. Then, we simulated the scenario where we manually inspect a fraction of training examples, correcting them if they were mislabelled. We trained 2-hidden layer MLP with 1024 hidden units using SGD with a batch size of 128. We used the damping term of $\lambda = 0.001$ for PBRF and influence functions. For each PBRF and influence estimation, we measured the influence of removing a single training training example (self-influence scores~\citep{koh2017understanding,khanna2019interpreting}) to identify the top influential training examples.

We prioritized inspecting training examples that obtained high scores generated by the PBRF and influence functions. The results are summarized in Figure~\ref{fig:fix}. Using the self-influence score generated by the PBRF, it is possible to detect over 80\% of the mislabelled training examples by only examining 20\% of training examples. Similarly, as influence functions closely align with the PBRF, influence functions can provide an efficient tool to help fix mislabelled training examples. Both PBRF and influence functions outperform the baseline of randomly selecting a subset of training examples to inspect if there are any mislabelled training examples.

\end{document}